\documentclass[letterpaper]{article} % DO NOT CHANGE THIS
\usepackage[dvipsnames]{xcolor}

\usepackage{aaai24}  % DO NOT CHANGE THIS
\usepackage{times}  % DO NOT CHANGE THIS
\usepackage{helvet}  % DO NOT CHANGE THIS
\usepackage{courier}  % DO NOT CHANGE THIS
\usepackage[hyphens]{url}  % DO NOT CHANGE THIS
\usepackage{graphicx} % DO NOT CHANGE THIS
\usepackage{natbib}  % DO NOT CHANGE THIS AND DO NOT ADD ANY OPTIONS TO IT
\usepackage{caption} % DO NOT CHANGE THIS AND DO NOT ADD ANY OPTIONS TO IT
\usepackage{algorithm}
\usepackage[noend]{algorithmic}

\usepackage{newfloat}
\usepackage{listings}
\DeclareCaptionStyle{ruled}{labelfont=normalfont,labelsep=colon,strut=off} % DO NOT CHANGE THIS
\floatstyle{ruled}
\newfloat{listing}{tb}{lst}{}
\floatname{listing}{Listing}
\usepackage{todonotes}
\usepackage{arydshln}
\usepackage{array}
\usepackage{amsmath}
\usepackage{pifont}
\usepackage{multirow}
\usepackage{booktabs}
\usepackage{enumitem}
\usepackage{amsmath}
\usepackage{amsfonts}
\usepackage{subcaption}

\usepackage{highlight}
\renewcommand{\opacity}{50}

\newcommand{\concept}[1]{\textit{\texttt{#1}}}

\newcommand{\conceptbed}{\textsc{\textsc{ConceptBed}}}

\newcolumntype{L}{>{\centering\arraybackslash}m{10cm}}

\newcommand*\colourcheck[1]{%
  \expandafter\newcommand\csname #1check\endcsname{\textcolor{#1}{\ding{52}}}%
}
\colourcheck{blue}
\colourcheck{green}
\colourcheck{red}
\newcommand{\ccd}{{\fontfamily{qcr}\selectfont CCD}}

\DeclareMathOperator{\E}{\mathbb{E}}

\DeclareMathOperator{\metric}{\text{\ccd{}}}

\title{\conceptbed{}: Evaluating Concept Learning Abilities of Text-to-Image Diffusion Models}
\author{
    Maitreya Patel\textsuperscript{\rm 1}\thanks{Corresponding Author: {\tt maitreya.patel@asu.edu}}, 
    Tejas Gokhale\textsuperscript{\rm 2}, 
    Chitta Baral\textsuperscript{\rm 1}, 
    Yezhou Yang\textsuperscript{\rm 1}
}
\affiliations{
    \textsuperscript{\rm 1} Arizona State University \\
    \textsuperscript{\rm 2} University of Maryland Baltimore County\\
    }

\begin{document}

\maketitle

\begin{abstract}
The ability to understand visual concepts and replicate and compose these concepts from images is a central goal for computer vision.
Recent advances in text-to-image (T2I) models have lead to high definition and realistic image quality generation by learning from large databases of images and their descriptions.
However, the evaluation of T2I models has focused on photorealism and limited qualitative measures of visual understanding.
To quantify the ability of T2I models in learning and synthesizing novel visual concepts (\textit{a.k.a.} personalized T2I), we introduce \conceptbed{}, a large-scale dataset that consists of 284 unique visual concepts, and 33K composite text prompts.
Along with the dataset, we propose an evaluation metric, Concept Confidence Deviation ($\metric$), that uses the confidence of oracle concept classifiers to measure the alignment between concepts generated by T2I generators and concepts contained in target images.
We evaluate visual concepts that are either objects, attributes, or styles, and also evaluate four dimensions of compositionality: counting, attributes, relations, and actions.
Our human study shows that $\metric$ is highly correlated with human understanding of concepts.
Our results point to a trade-off between learning the concepts and preserving the compositionality which existing approaches struggle to overcome.
The data, code, and interactive demo is available at: \url{https://conceptbed.github.io/}
\end{abstract}

\section{Introduction}

\begin{figure}[t]
  \begin{center}
    \includegraphics[width=\linewidth]{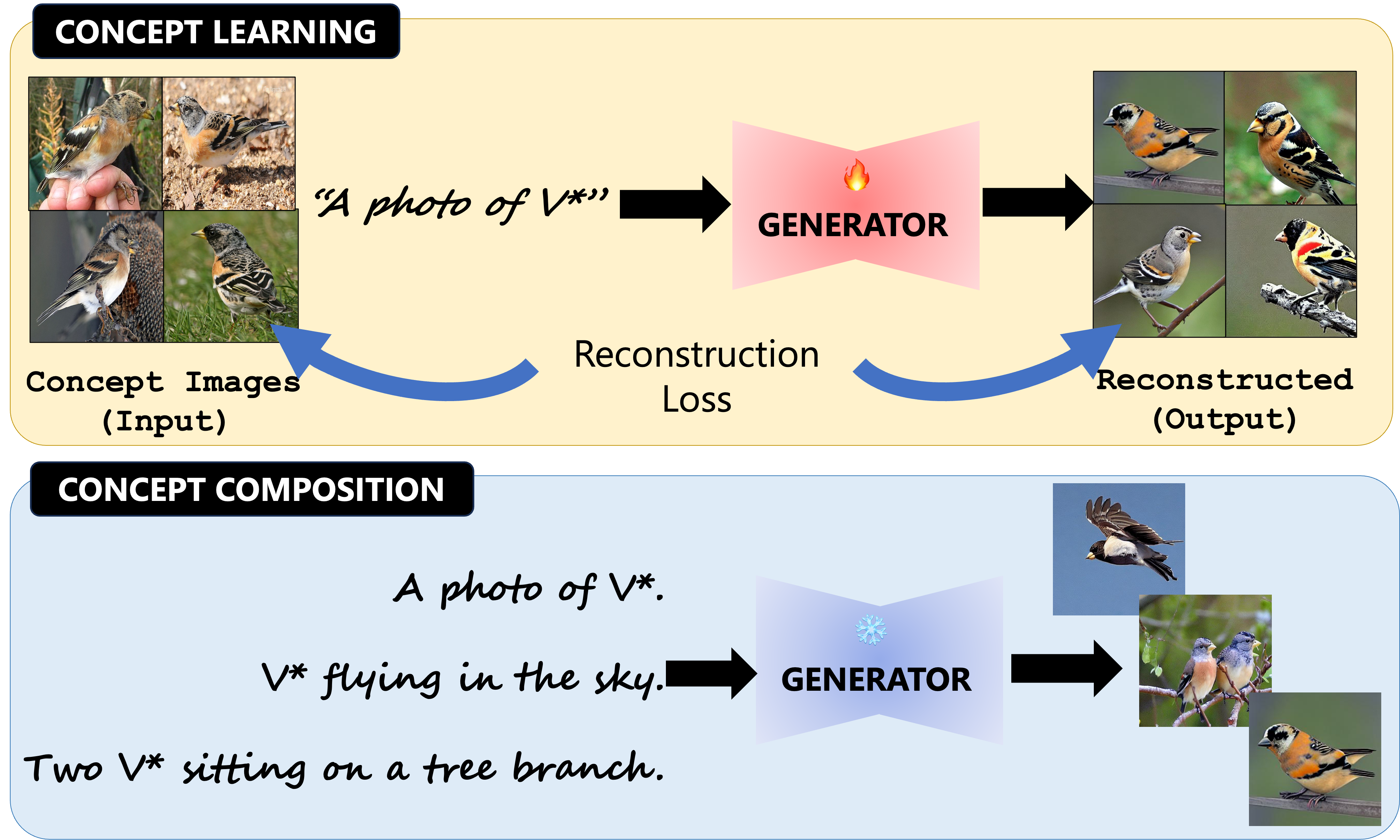}
  \end{center}
  \caption{
  Visual concept learners such as textual inversion models learn to \textit{``invert''} a set of images (about a concept $c$) into a text embedding $\mathrm{V}^\ast$, and then use this learned textual concept in new text prompts to generate images of concept $c$ under different contexts and by performing novel compositions with other concepts.
  The proposed \conceptbed{} dataset along with the evaluation metric $\metric$ allows us to comprehensively and quantifiably evaluate concept learning abilities of text-to-image diffusion models.
  } 
  \label{fig:introduction}
\end{figure}

Humans reason about the visual world by aggregating entities that they see into “visual concepts”: both \concept{cats} and \concept{elephants} are \concept{animals}, and both \concept{palms} and \concept{pines} are \concept{trees}.
We use natural language to describe images and things that we see.
Although this type of visual concept learning is well-defined in human psychology~\cite{murphy2004big}, it remains elusive in the context of data-driven techniques capable of learning and reasoning from images and their natural language descriptions.

Text-to-Image (T2I) generative models are trained to translate natural language phrases into images that correspond to that input.
High-quality T2I models, therefore, serve as a link between human-level concepts (expressed in natural language) and their visual representations and are one way to reproduce visual concepts.
On the other hand, this has also sparked interest in visual concept learning (\textit{a.k.a.} personalized T2I) through the procedure of \textit{``image inversion''} -- to translate one or many images corresponding to a visual concept into a latent representation of that visual concept.
While earlier methods primarily explored image inversion using generative adversarial networks \cite{xia2022gan}, 
methods such as Textual Inversion \cite{gal2022image} and Dreambooth \cite{ruiz2022dreambooth} combine image inversion with T2I
-- this has led to an effective way to quickly learn concepts from a few images and reproduce them in novel combinations and compositions with other concepts, attributes, styles, etc.
These methods aim to learn concepts with minimal reference images by fine-tuning pre-trained text-conditioned diffusion models (Figure~\ref{fig:introduction}).
Therefore this paradigm of T2I and image inversion is a powerful new way of learning and reproducing concepts.

Within this paradigm of novel visual concept learning via image inversion, two primary evaluation criteria have emerged: (1) concept alignment, which assesses the correspondence between the generated images and the target concept images, and (2) composition alignment, which evaluates whether the generated images maintain compositionality. 
Previous studies have been small scale, evaluating only a small number of hand-picked concepts and compositions; as such making generic claims via such findings is difficult.
Furthermore, the established evaluation metrics  such as DINO-based cosine similarity \cite{ruiz2022dreambooth} (for measuring concept alignment), KID \cite{kumari2022multi} (for measuring the amount of concept overfitting), and CLIPScore \cite{hessel2021clipscore} (for evaluating compositionality), have encountered challenges in accurately capturing human preferences.
Consequently, there is a growing need for better automated evaluations.

Therefore, we introduce \conceptbed{}, comprehensive dataset and evaluation framework that is aligned with human preferences.
% Therefore, we introduce the \conceptbed{} benchmark: a comprehensive evaluation strategy aligned with human preferences and accompanied by the large concept dataset. 
The \conceptbed{} dataset comprises 284 distinct concepts and approximately 33,000 composite text prompts, which can be further extended using the provided automatic realistic dataset creation pipeline.
The dataset focuses on four diverse concept learning evaluation tasks: learning styles, learning objects, learning attributes, and compositional reasoning. 
To gain a deeper understanding of previous methodologies, we incorporate four composition categories -- action, attribution, counting, and relations.

% Armed with the dataset, we show how it can be used to evaluate concept learners.
% To that end, we present a novel concept deviation-aware evaluation framework that exhibits a strong correlation with human preferences. 
We use our large-scale dataset to evaluate concept learners, by developing a novel evaluation metric called Concept Confidence Deviation (\ccd{}). We conduct a human study and find that relative evaluations of models in terms of \ccd{} are well aligned with human preferences.
Therefore, \ccd{} combined with the \conceptbed{} dataset, offers an alternative to existing evaluation strategies, facilitating more effective large-scale evaluations.
For each evaluation criteria, we train supervised classifiers (oracles) to detect whether generated concept images are accurate.
% For each of the four tasks within \conceptbed{}, we train a supervised classifier (referred to as an oracle) to detect the respective concepts. 
Subsequently, the confidence scores from these oracles are utilized to calculate the instance-level concept deviations of the generated concept images in relation to the reference target ground truth images using the proposed $\metric$ metric.
This approach enables us to assess concept and composition alignment more effectively.
We further show that \ccd{} calculated using a pre-trained few-shot classifier also maintains a high correlation with human preferences.
This allows \ccd{} to measure concept alignment on unseen concepts.

\begin{figure}
    \centering
    \includegraphics[width=\linewidth]{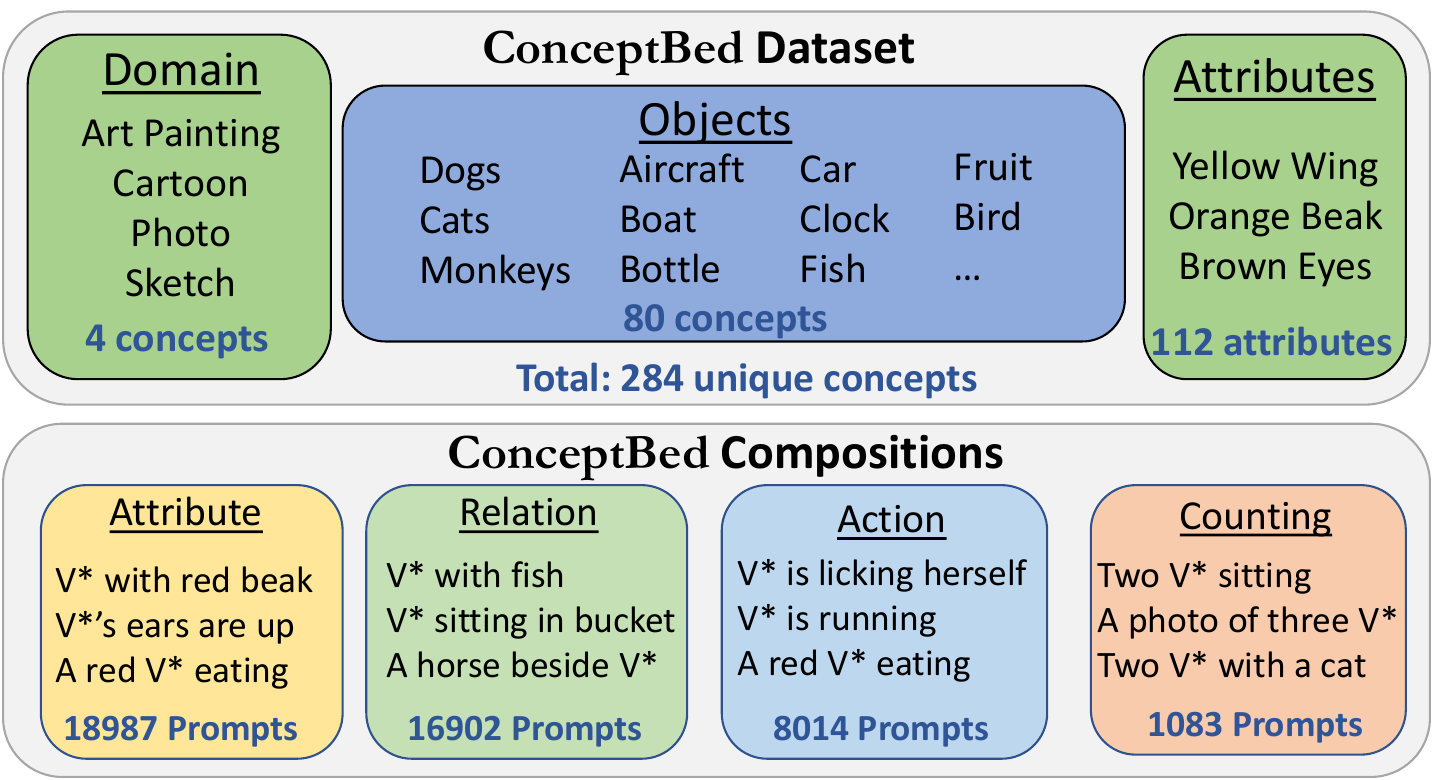}
    \caption{
    A summary of the \conceptbed{} dataset for large-scale grounded evaluations of concept learners. 
    The collection of concepts is categorized into three classes: (1) Domain, (2) Objects, and (3) Attributes. 
    \conceptbed{} has 284 unique concepts and four compositional categories.
    Here, V* is a learned concept.
    }
    \label{fig:teaser}%
\end{figure}

We conduct extensive experiments on four recently proposed concept learning methodologies. 
In total, we fine-tune approximately 1100 models (one model per concept) and generate over 500,000 concept-specific images. 
Our results reveal a surprising trade-off between concept alignment and composition alignment, wherein methods excelling at concept alignment tend to fall short in preserving compositions and vice versa. 
This suggests that previous concept learning approaches are either highly overfitted or severely underfitted. 
Furthermore, our experiments demonstrate that utilizing a pre-trained CLIP~\cite{radford2021learning} textual encoder aids in maintaining compositionality, but it lacks the flexibility required to learn complex concepts, such as \concept{sketch}.

In summary, we make the following \textbf{key contributions:}
\begin{itemize}[nosep,noitemsep,leftmargin=*]
    \item We introduce \conceptbed{}, a comprehensive benchmark for grounded quantitative evaluations of text-conditioned concept learners. 
    % It comprises 284 unique and diverse concepts and approximately 33,000 composite text prompts across four composition categories.
    \item The \underline{C}oncept \underline{C}onfidence \underline{D}eviation ($\metric$) evaluation metric, measures the learners' ability to preserve concepts and compositions. We demonstrate a strong correlation between $\metric$ and human preferences.
    \item Through extensive experiments with 1,100+ models, we identify shortcomings in prior works and suggest future research directions. \conceptbed{} sets a standard for evaluating personalized text-to-image generative models.
\end{itemize}

\section{Preliminaries}
\label{sec:t2i_paradigm}

Prior studies on concept learning have focused on text-conditioned diffusion models, such as Textual Inversion~\cite{gal2022image}, DreamBooth~\cite{ruiz2022dreambooth}, and Custom Diffusion~\cite{kumari2022multi}. 
These models operate within the T2I paradigm, where a text prompt ($y$) serves as input to generate the corresponding image ($x_{gen}$) representing the given prompt $y$. 
A popular approach within T2I is the Latent Diffusion Model (LDM)~\cite{rombach2022high}, which incorporates two key modules:
\begin{enumerate}[noitemsep]
    \item Textual Encoder ($C_\theta$): This module generates embeddings corresponding to the input text prompt;
    \item Generator ($\epsilon_\phi$): The generator estimates the noise iteratively from the input randomly sampled matrix at timestamp $t$ ($z_t$), conditioned on the text.
\end{enumerate}

Since T2I models solely consider text input, the target concept ($c$) is represented in terms of text tokens. 
These tokens can subsequently be employed to generate images associated with concept $c$. 
Therefore, in Textual Inversion, the concept learning task is approached as an image inversion problem, aiming to map the target concept back to the text-embedding space. 

Let V\textsuperscript{*} denote the text tokens corresponding to the learned concept $c$. Once the optimal mapping from V\textsuperscript{*} to the target concept is determined, we can generate concept-specific images using the LDM by providing V\textsuperscript{*} in the text prompt. Suppose we are provided with $m$ images ($X_{1:m}$) of the target concept $c$.
Now, in order to learn the text tokens V\textsuperscript{*} corresponding to the concept $c$ from the set of images $X_{1:m}$, the Textual Inversion methodology aims to optimize V\textsuperscript{*} by reconstructing $X_{1:m}$ using the objective function of the LDM with frozen parameters $\theta$ and $\phi$:

\begin{equation}
    \small
    V^\ast = \underset{v}{\mathrm{argmin}}
    % \mathcal{L}_{LDM} = 
    \underset{
        \substack{
            x \in X_{1:m}, ~t, \\
            \epsilon \sim \mathcal{N}(0,1), ~z \sim \mathcal{E}(x)
            }
        }{\mathbb{E}} ||\epsilon -  \epsilon_{\phi}(z_t,t,x,C_\theta(y))||_{2}^{2}
\end{equation}

In the case of DreamBooth and Custom Diffusion, instead of finding the optimal V\textsuperscript{*}, it optimizes the model parameter $\phi$ associated with the noise estimator ($\epsilon_{\phi}$). 
This optimization process enables the model to learn the mapping between randomly initialized V\textsuperscript{*} and the target concept $c$. 

\begin{equation}
    \small
    \phi^\ast = \underset{\phi}{\mathrm{argmin}}
    \underset{
        \substack{
            x \in X_{1:m}, ~t, \\
            \epsilon \sim \mathcal{N}(0,1), ~z \sim \mathcal{E}(x)
            }
        }{\mathbb{E}} ||\epsilon -  \epsilon_{\theta}(z_t,t,x,C_\phi(y))||_{2}^{2}
\end{equation}

\noindent Once $\phi^{*}$ is obtained, it can be used to generate images related to the target concept.\footnote{DreamBooth and Custom-Diffusion use additional regularizer to improve compositionally by using same objective function on a diverse set of image-caption pairs.}

Once the images are generated, in order to evaluate these generated images, it is essential to verify whether they align with the learned concepts while maintaining compositionality. 

\begin{figure}[t]
    \centering
    \includegraphics[width=\linewidth]{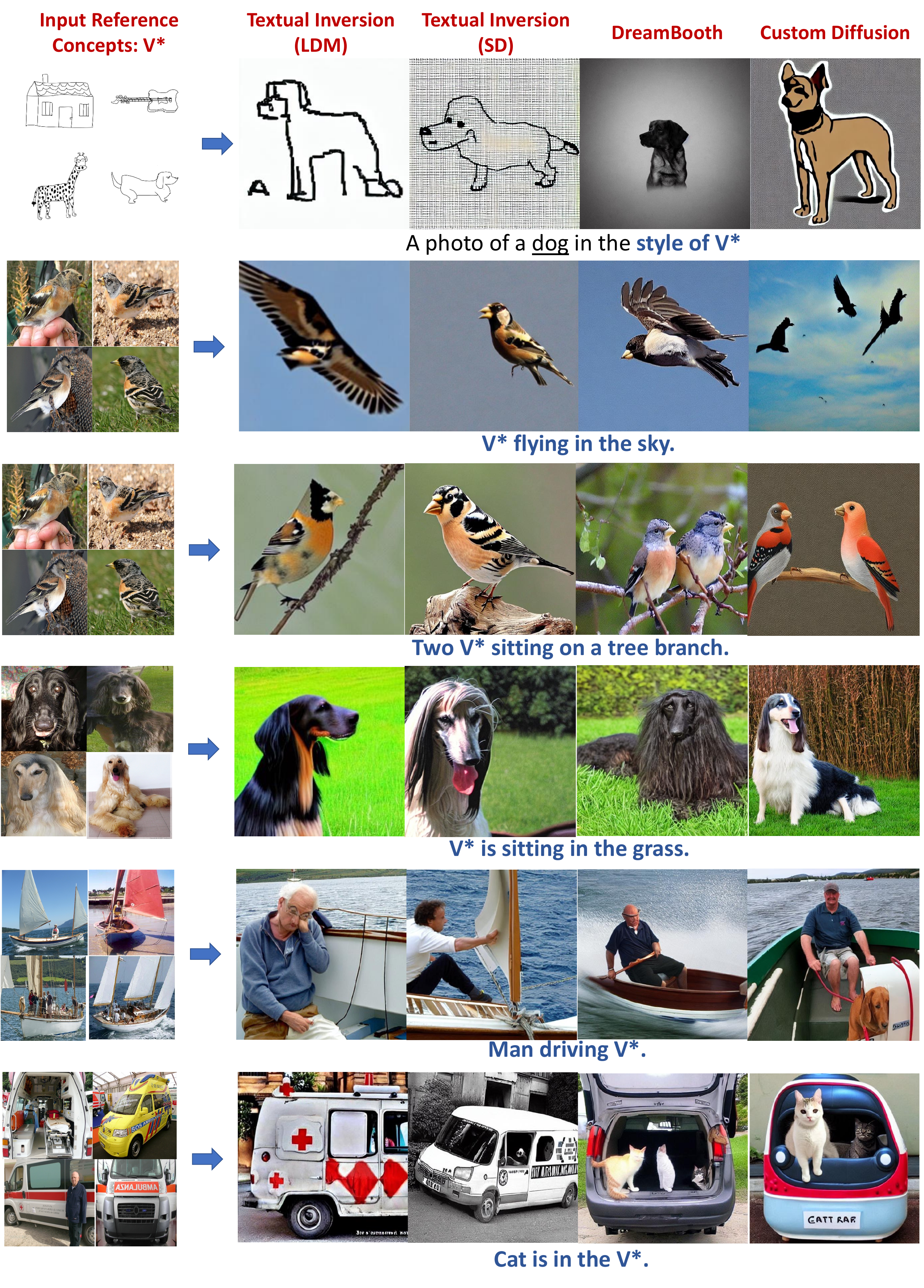}
    \caption{Qualitative examples showcasing the effectiveness of concept learners on the \conceptbed{} dataset. The leftmost column displays four instances of ground truth target concept images (V\textsuperscript{*}). Subsequent columns exhibit target concept-specific images generated by all baseline methods.
    }
    \label{fig:qualitative_examples}%
\end{figure}

\begin{algorithm}[t]
    \caption{Concept Confidence Deviation}
    \label{alg:eval_algo}
    \small
    \begin{algorithmic}[0]
        % \REQUIRE $n \geq 0 \vee x \neq 0$
        \STATE \textbf{Input:} Concept fine-tuned models $G \in \{g_c\}$, $c \in C_{\conceptbed{}}$;  \\
        \quad Oracles $F_t \in \{F_{PAC}, F_{Imagenet}, F_{CUBS}, F_{VQA}\}$; \\
        \quad Reference set of concept images $X^{ref} \in \{x_c\}$; \\
        \quad Target set of prompts $Y \in \{y_c\}$; \\
        % \quad Task $t \in \{\mathrm{ImageNet}, \mathrm{PAC}, \mathrm{CUBS}, \mathrm{Composition}\}$;
        \STATE \textbf{Output:} Estimated $\metric$
    \end{algorithmic}
    \begin{algorithmic}[1]
        \STATE \textbf{Initialize:} $score = []$; $p^{real} = []$
        \FOR{$c \in C_{\conceptbed{}}$}
            \STATE $p^{real} = []$
            \IF{$t = VQA$}
                \STATE $c = 3$
            \ENDIF
            
            \FOR{$x=1 \dots M$}
                \STATE $p^{real} \gets F_t(x_i, c)$
            \ENDFOR
            \STATE $\Bar{p}^{real} = \frac{1}{M} \sum_{i=1}^{M} p^{real}_{i}$
            \FOR{$n=1 \dots N$}
                \STATE $x_{gen} = g_c(y_c)$
                \STATE $score \gets -1 * (F(x_{gen}, c) - \Bar{p}^{real})$  \hfill\COMMENT{// Eq. \ref{eq:ccd}}
            \ENDFOR
        \ENDFOR
        \STATE $\metric = \frac{1}{NC} \sum_{i=1}^{NC} score_{i} $
    \end{algorithmic}
    
\end{algorithm}

\section{\conceptbed{}}
In this section, we introduce \conceptbed{}, a comprehensive collection of concepts, designed to accurately estimate concept and composition alignment by quantifying deviations in the generated images. 
Later, we introduce the novel evaluation framework associated with \conceptbed{}.
Please refer to the Appendix for additional insights on the proposed dataset and evaluation framework.
% \footnote{Please refer to the arxiv release for additional insights on the proposed dataset and evaluation framework: \url{https://arxiv.org/abs/2306.04695}.}

\subsection{\conceptbed{}: Dataset Construction}
\label{sec:dataset}

\conceptbed{} 
incorporates existing datasets such as ImageNet~\cite{deng2009imagenet}, PACS~\cite{li2017deeper}, CUB~\cite{wah2011caltech}, and Visual Genome~\cite{krishna2017visual}, enabling the creation of a labeled dataset.
Figure \ref{fig:teaser} provides an overview of the \conceptbed{} dataset.

\smallskip\noindent \textbf{Learning Styles.} We use styles from the PACS dataset: \concept{Art Painting}, \concept{Cartoon}, \concept{Photo}, and \concept{Sketch}. 
Each style contains images corresponding to seven categories.
The concept learner aims to use examples from one style as a reference and generate style-specific images for all seven entities.

\smallskip\noindent\textbf{Learning Objects.} Extracting object-level concepts is accomplished through the utilization of the ImageNet dataset. 
It comprises 1000 low-level concepts from the WordNet~\cite{fellbaum2010wordnet} hierarchy. 
However, due to the presence of noise in ImageNet images and the lack of relevance to daily life for many concepts, we employ an automated filtering pipeline to ensure the usefulness and quality of the reference concept images. 
The pipeline involves extracting a list of low-level concepts and their parent concepts from ImageNet, followed by extracting text phrases from Visual Genome containing the concept as a subject in the caption. 
If an insufficient number of such captions exists (less than 10 in Visual Genome) or they cannot be found, the concepts are discarded. 
This filtering process results in 80 concepts such as (\concept{brambling}, \concept{squirrel monkey}, etc.). 
We select the top 100 high-quality images for each concept that will be used to train the concept learning methodologies.

\smallskip\noindent\textbf{Learning Attributes.} Since ImageNet dataset images are not labeled based on the attributes present in the image, it is necessary to rely on datasets that provide attribute-level grounded labels. 
Therefore, we additionally employ the CUB dataset, which offers attribute-level labels (such as \concept{orange wing}, \concept{blue forehead}, etc.), enabling the \conceptbed{} to perform evaluations and measure the attribute-level performance of concept learners. 

\begin{figure}
    \centering
    \includegraphics[width=\linewidth]{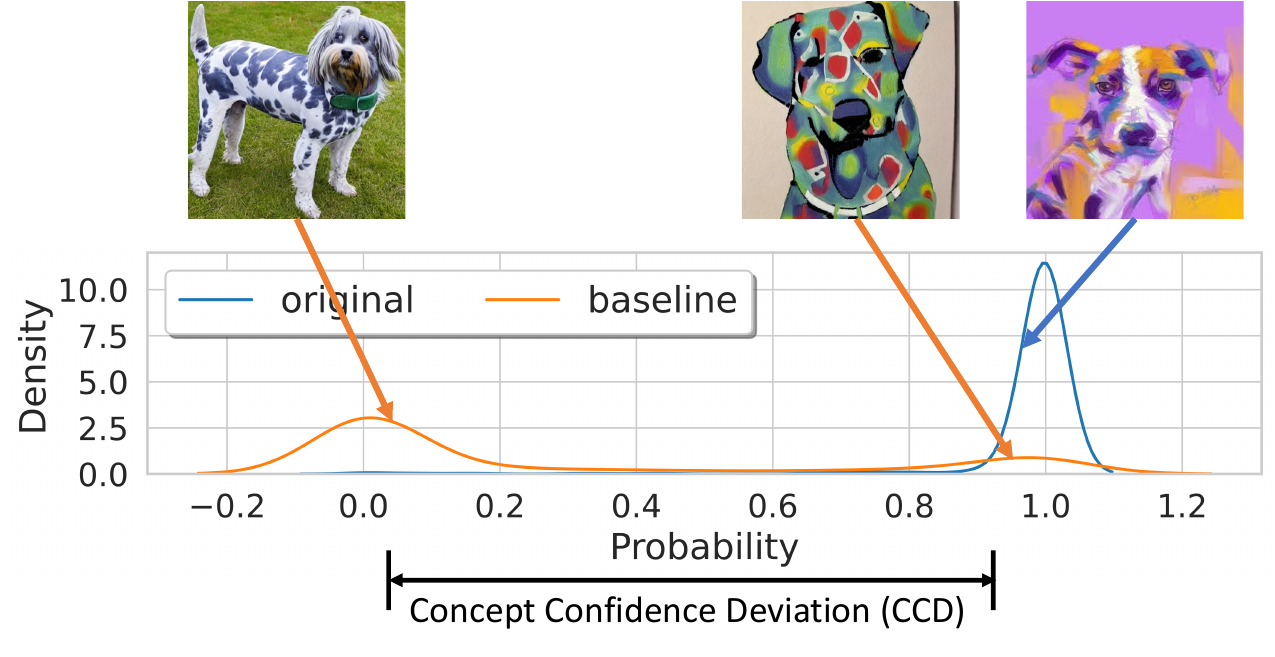}
    \caption{
    Intuitive illustration of the Concept Confidence Deviation (\ccd{}) for the concept \concept{Art Painting}. {Blue} and {Orange} are the probability distributions of the real and generated concept images.
    }
    \label{fig:ccd_intuit}%
\end{figure}

\smallskip\noindent\textbf{Compositional Reasoning.} 
In addition to learning new concepts, it is crucial to maintain prior knowledge and associate the acquired concepts with it. 
To conduct these evaluations holistically, we use Visual Genome to extract captions in which the concept appears as the subject of the sentence. 
These captions are categorized into four composition categories (actions, attributes, counting, and relation) through few-shot classification using GPT3~\cite{brown2020language}. 
This categorization allows us to measure the performance of the baselines on each category, and an in-depth understanding of the varying difficulty levels of different compositions.

\subsection{\conceptbed{}: Dataset Statistics}
The \conceptbed{} dataset consists of 284 unique concepts, comprising 80 concepts from ImageNet, 200 concepts from CUB, and 4 concepts from PACS. 
In total, the dataset contains approximately 33,000 composite prompts for the evaluation of all 80 processed concepts from ImageNet, with each composite prompt having up to two composition categories. 
Out of these composite prompts, 18987, 16902, 8014, and 1083 prompts contribute to the attribute, relation, action, and counting categories, respectively. 

Our dataset curation pipeline is flexible to be extended to larger datasets such as OpenImages-v7~\cite{kuznetsova2020open} and LAION-5B~\cite{schuhmannlaion}. 
However, it is important to note that this extension would significantly increase the resource requirements. 
With the introduction of this dataset, our primary objective is to provide a standardized and benchmarked evaluation framework for concept learners, enhancing research in the field.
% Further details on dataset creation and insights are in the appendix.

\subsection{$\metric$: Concept Confidence Deviation}

\label{sec:framework}

\begin{table*}[t]
\centering
% \small
    \fontsize{9pt}{9pt}
    \selectfont

\begin{tabular}{l cc c cc c cc}
\toprule
\multirow{2}{*}{\textbf{Model}} &  \multicolumn{2}{c}{\textbf{\textbf{Concepts}}} & \hphantom & \multicolumn{2}{c}{\textbf{Fine-grained\textsubscript{CUB}}} & \hphantom & \multirow{2}{*}{\textbf{Composition}} \\
\cmidrule{2-3}
\cmidrule{5-6}
  &  \textbf{Domain\textsubscript{PACS}} & \textbf{Objects\textsubscript{ImageNet}} & \hphantom & \textbf{Object-level} & \textbf{Attribute-level} & \hphantom &  \\
  
\midrule

TI (LDM) & \textbf{0.0478}           & 0.0955           && 0.2289  & 0.1174   && 0.1906\\
TI (SD)  & 0.2456 & \textbf{0.0472}  && \textbf{0.0859}   & \textbf{0.0332}    && \textbf{0.1090}\\
DB              & \underline{0.6825}  & 0.0678           && 0.0963            & 0.0469             && 0.3527\\
CD        & 0.6206           & \underline{0.2085} && \underline{0.3934}            & \underline{0.1743}             && \underline{0.4916}\\
\midrule
Original                &  0.0000                     & 0.0000                      && 0.0000                       & 0.000                         && 0.0000\\
\bottomrule
\end{tabular}%
\caption{
Results of Concept Alignment Evaluation.
The table shows the performance of concept learners evaluated using the $\metric$ ($\downarrow$) metric for Concepts (Domain\textsubscript{PACS}, and Object\textsubscript{ImageNet}), Fine-grained\textsubscript{CUB} (Object-level, and Attribute-level), and Composition.
The best and worst performing models are indicated by bold and underlined numbers, respectively.
}
\label{tab:main_results}

\end{table*}
\begin{table*}[t]
    \centering
        \fontsize{9pt}{9pt}
    \selectfont

\setlength\tabcolsep{4.5pt}
    % \small
    % \resizebox{\textwidth}{!}{%
        \begin{tabular}{@{}l ccc c ccc c ccc c ccc@{}}
            \toprule
            \multirow{2}{*}{\textbf{Models}} & \multicolumn{3}{c}{\textbf{Relation}} &\phantom{}& \multicolumn{3}{c}{\textbf{Action}} &\phantom{}& \multicolumn{3}{c}{\textbf{Attribute}} &\phantom{}& \multicolumn{3}{c}{\textbf{Counting}} \\

             & \textbf{CLIP  } & \textbf{VQA.} & \textbf{$\metric$ } && \textbf{CLIP  } & \textbf{VQA} & \textbf{$\metric$ } &&  \textbf{CLIP  } & \textbf{VQA.} & \textbf{$\metric$ } && \textbf{CLIP  } & \textbf{VQA.} & \textbf{$\metric$ } \\
            \midrule

             TI (LDM) & 0.6589 & 66.60\% & 0.2074 && 0.6523 & 68.69\% & 0.2098 && 0.6599 & 72.22\% & 0.1331 && 0.6515 & 65.78\% & 0.1231 \\
        TI (SD) & 0.6294 & 70.09\% & 0.1735 && 0.6274 & 70.81\% & 0.1884 && \underline{0.6360} & 74.75\% & 0.1091 && 0.6301 & 68.38\% & 0.1020 \\ 
        DB & \underline{0.7051} & 82.20\% & 0.0542 && \underline{0.6995} & 84.61\% & 0.0496 && \underline{0.6862} & 82.24\% & 0.0355 && \underline{\textbf{0.6924}} & \underline{78.90\%} & \underline{-0.0016} \\ 
        CD & \underline{\textbf{0.7065}} & \textbf{82.94\%} & \textbf{0.0471} && \underline{\textbf{0.7053}} & \textbf{86.35\%} & \textbf{0.0347} && \underline{\textbf{0.6940}} & \textbf{84.20\%} & \textbf{0.0163} && \underline{0.6921} & \underline{\textbf{79.36\%}} & \underline{\textbf{-0.0054}} \\ 

        \midrule
        
        SD & \underline{0.7222} & 83.42\% & 0.0403 && \underline{0.7178} & 87.39\% & 0.0256 && \underline{0.7053} & 83.85\% & 0.0184 && \underline{0.7085} & \underline{81.07\%} & \underline{-0.0206} \\ 
        Original & 0.6626 & 87.45\% & 0.0000 && 0.6831 & 89.78\% & 0.0000 && 0.6306 & 85.79\% & 0.0000 && 0.6553 & 78.32\% & 0.0000 \\ 

        \bottomrule

        \end{tabular}
    % }
\caption{
Compositional Reasoning Evaluation Results.
The table shows the performance of the prior works for Composition Alignment. 
CLIP $(\uparrow)$ is the traditional image-text alignment metric.
VQA $(\uparrow)$ is the accuracy of the ViLT VQA classifier on generated boolean questions. 
And $\metric$ ($\downarrow$) is the composition deviations reported from the ViLT model with respect to its performance on original images.
The best-performing model is indicated by bold numbers, while the performance that is higher than the original data is reported with underline.
}
\label{tab:composition_results}
\end{table*}

\begin{table*}[t]
    \centering
        \fontsize{9pt}{9pt}
    \selectfont
\setlength\tabcolsep{4pt}

    % \Huge
    % \resizebox{\textwidth}{!}{%
    \begin{tabular}{@{}l cccc c cccc c ccc@{}}
        \toprule
        \multirow{2}{*}{\textbf{Models}} & \multicolumn{4}{c}{\textbf{Domain\textsubscript{PACS}}} & \phantom{} & \multicolumn{4}{c}{\textbf{Objects\textsubscript{ImageNet}}} & \phantom{} & \multicolumn{3}{c}{\textbf{Compositional Reasoning}}  \\
        \cmidrule{2-5}
        \cmidrule{7-10}
        \cmidrule{12-14}

        & \textbf{DINO ($\uparrow$)} & \textbf{KID ($\downarrow$)} &\textbf{ $\metric$ ($\downarrow$)} & \textbf{H.S. ($\uparrow$)}  && \textbf{DINO ($\uparrow$)} & \textbf{KID ($\downarrow$)} &\textbf{ $\metric$ ($\downarrow$)} & \textbf{H.S. ($\uparrow$)}  && \textbf{CLIP ($\uparrow$)} &\textbf{ $\metric$ ($\downarrow$)} & \textbf{H.S. ($\uparrow$)}  \\
        \midrule

        TI (LDM)                                  &    0.5073 & 0.0117 & 0.0478   & 4.028 &  & 0.4708  & 0.0552 & 0.0955   & 4.069 &  & 0.6611    & 0.1684    & 2.851   \\
        TI (SD)                                    &   0.4104 & 0.0422  & 0.2456   & 4.084 &  & 0.4457  & 0.0294 & 0.0472   & 4.159 &  & 0.6309    & 0.1432    & 3.694   \\
        DB                                         & 0.3925 & 0.1101  & 0.6825   & 3.083 &  & 0.4525  & 0.0290 &  0.0678   & 4.075 &  & 0.6919    & 0.0344    & 3.556   \\
        CD                                         &   0.3956  & 0.0593 & 0.6206   & 3.164 &  & 0.4450  & 0.0492 & 0.2085   & 3.803 &  & 0.6968    & 0.0232    & 4.178   \\
        \midrule
        Correlation                     &    0.6557  & \underline{-0.8252}   & \textbf{-0.9515}   & 1.000   &  & 0.2787 & \underline{-0.5347} & \textbf{-0.9892}  & 1.000   &  & \underline{0.3486}    & \textbf{-0.7342}   & 1.000    \\

        \bottomrule
    \end{tabular}%
    % }
    \caption{
    Human Evaluations.
    Comparison of prior quantitative metrics and $\metric$ metric with Human evaluations. 
    DINO based pairwise cosine similarity is the prior evaluation metric~\cite{ruiz2022dreambooth}.
    KID was used to measure the overfitting by \cite{kumari2022multi}.
    CLIP (CLIPScore) is the traditional reference-free image-text similarity metric.
    $\metric$ is our presented concept deviation-aware evaluation metric.
    H.S. denotes the corresponding Human Score.
    Here, Domain\textsubscript{PACS} and Object\textsubscript{ImageNet} evaluations are for concept alignment and composition alignment is for image-text similarity.
    A high negative correlation between \ccd{} and human ratings implies strong alignment, as lower \ccd{} and higher human ratings correspond to better performance.
    }
    \label{tab:human_eval}
\end{table*}

\begin{table}[]
    \centering
    \fontsize{9pt}{9pt}
    \selectfont

    % \small
    % \resizebox{\linewidth}{!}{%
    \begin{tabular}{l | ccc}
        \toprule
        \multirow{2}{*}{\textbf{Model}} & \textbf{PACS} & \multicolumn{2}{c}{\textbf{ImageNet}} \\
         & \textbf{Domain} & \textbf{Object} & \textbf{Composition} \\
        \midrule
        \textbf{TI (LDM)} & 72.84 & 64.53 & 58.28 \\ 
        \textbf{TI (SD)} & 52.25 & 70.79 & 65.42 \\ 
        \textbf{DB} & 24.71 & 67.45 & 39.42 \\ 
        \textbf{CD} & 20.12 & 52.06 & 26.31 \\ \bottomrule

    \end{tabular}
    % }
    \caption{
Recall.
Percentage of generated images highly aligned ($\metric <= 0.0$) with the target concept images.
}
\label{tab:recall}

\end{table}

\subsubsection{Problem Statement.}
Consider a pre-trained text-conditioned diffusion model $g(\cdot)$, which can be further fine-tuned on a specific concept $c$ such that $c \in \mathcal{C}_{\conceptbed{}}$. 
We assume the availability of concept-specific target images from the \conceptbed{} dataset, denoted as $\mathcal{D}_c^{real} {\in} \mathcal{D}_{\conceptbed{}}^{test}$. 
Denote the concept learner $g(\cdot)$ fine-tuned on concept $c$ using $\mathcal{D}_c^{real}$ as $g_c(\cdot)$. 
First, we generate a collection of $N$ images using the learned concept $c$,
and denote this set of images as 
$\mathcal{D}_c^{gen} {=} \{ x_i^{gen} {=} g_c(p_c^i, s^i);~ \forall i \in [0, N]\}$, where $p_c^i$ is the concept-specific prompt and $s$ is the random seed. 

The alignment between two distributions (i.e., $D_c^{real}$ and $D_c^{gen}$) is typically computed by first extracting features from the model $m$ (i.e., $f_{real} {=} m(D^{real})$; $f_{gen} {=} m(D^{gen})$) and then employing a distance metric $d$ (i.e., $score = d(f_{real}, f_{gen})$). 
Several combinations of models ($m$) and distance measures ($d$) have been used in prior work.
For concept alignment, \citet{ruiz2022dreambooth} use $m{=}\mathrm{DINO}$ with $d{=}\mathrm{Cos}$ and \citet{kumari2022multi} use $m{=}\mathrm{Inception}$ with $d{=}\mathrm{KID}$.
For composition alignment, all prior work utilizes $m{=}\mathrm{CLIP}$ with $d{=}\mathrm{Cos}$.
However, these methods fail to accurately capture the concept deviations within the generated images; rendering them ineffective in comparing performance across the methodologies (as shown in Section~\ref{sec:results}).

\subsubsection{Concept Confidence Deviation (\ccd{}).}
\label{sec:eval_metric}

To address the above limitations, we propose training the oracle classifier $F$, specifically for the concept detection task using the \conceptbed{} training dataset, $\mathcal{D}_{\conceptbed{}}^{train}$.
Then one can simply use $m=F$ and $d=\mathrm{Accuracy}$ to verify whether $x_{gen}$ is aligned with $x_{real}$.
However, measuring accuracy does not allow instance-level evaluations.
By leveraging the output probabilities of the oracle (concerning the concept label $y_c$), we can estimate the deviations associated with each generated image $x_{gen}$ w.r.t.\ the output probabilities of real target images $x_{real}$.
\underline{C}oncept \underline{C}onfidence \underline{D}eviation is defined as:

\begin{equation}
    \small
    \metric = - \underset{c}{\E} \bigg [ \underset{x_{gen}}{\E} \big [ F(y_c|x_{gen}) - \underset{x_{real}}{\E} F(y_c|x_{real}) \big] \bigg ].
    \label{eq:ccd}
\end{equation}

$\metric$ first calculates the mean target probability on the test ground truth images and then measures the difference in probability of the generated images. 
$\metric$ with negative or close to $0.0$ values indicates that the generated images closely follow the distribution of the ground truth concept images. 
A positive $\metric$ value suggests that the generated images deviate from the original distribution.
Figure~\ref{fig:ccd_intuit} shows an intuitive example of \ccd{} by calculating the distance between two probability densities corresponding to the real and generated target concept.

\subsection{Task Specific Evaluation Settings}
\label{sec:eval_setting}

To efficiently leverage the \conceptbed{} evaluation pipeline, we trained separate oracles on the corresponding \conceptbed{} datasets. Two different types of evaluations are conducted, each with its respective set of oracles:
1) concept alignment, measured by concept classifiers, and
2) compositional reasoning, measured by a VQA model.

\subsubsection{Concept Alignment:} Concept alignment evaluation was performed on all tasks, including the generated concept images with different composite text prompts. 
To evaluate the style, a ResNet18~\cite{he2015deep} model is trained to distinguish the images between four style concepts. 
To evaluate the objects, a ConvNeXt~\cite{liu2022convnet} model is fine-tuned on 80 classes from the \conceptbed{} using the ImageNet training subset. 
The Concept Embedding Model (CEM)~\cite{zarlenga2022concept} was trained on CUB to detect the concepts and attributes. 
Images corresponding to the concepts were generated for each task by following the prompts: ``A photo of V\textsuperscript{*}" for objects and ``A photo of a \textit{$<$entity-name$>$} in the style of V\textsuperscript{*}" for styles. 
Here, \textit{$<$entity-name$>$} belongs to the seven classes from {PACS}.
The remaining task, composition, utilizes the same pre-trained ConvNeXt model for concept alignment, as \conceptbed{} compositions are specifically for 80 ImageNet concepts.

\subsubsection{Compositional Reasoning:} To measure the image-text alignment with respect to the input prompts, the concept-specific token ($\mathrm{V\textsuperscript{*}}$) was removed and replaced with the corresponding ground truth label (i.e., \concept{dogs}, \concept{cats}, etc.). 
The image-text similarity was then measured. 
Unlike previous works, CLIP was not used due to its inability to capture compositions~\cite{thrush2022winoground}. 
Instead, taking after~\cite{cho2022dall}, we propose to use a pre-trained ViLT~\cite{kim2021vilt} as a VQA model for composition evaluations.
Specifically, from each composite prompt, the boolean questions with positive answers are generated~\cite{banerjee2021weaqa}.
As ViLT is essentially a classifier, the $\metric$ can be calculated with respect to the confidence of the model associated with a ``yes" answer.
\section{Experiments \& Results}
In this section, we benchmark four state-of-the-art concept learning methodologies.
We first explain the experimental setup and report the evaluation results using the \conceptbed{} framework along with human preferences.
Additional details about the experimental setup, results, and human evaluations are in the appendix.
% \footnote{The arxiv release contains additional details about the experimental setup, results, and human evaluations: \url{https://arxiv.org/abs/2306.04695}}

\subsection{Experimental Setup}
\label{sec:exsetup}

In our experiments, we study four text-conditioned diffusion modeling-based concept learning strategies: Textual Inversion (TI) on LDM and SD, DreamBooth (DB)~\cite{ruiz2022dreambooth}, and Custom Diffusion (CD)~\cite{kumari2022multi}.
We generate $N=100$ images for all concepts to measure the concept alignment and $N=3$ images for 33K composite text prompts.
For a total of 284 concepts, we train all four baselines. 
This leads to $1100$+ concept-specific fine-tuned models and we generate a total of $500,000$ images
% \footnote{284*100*4 + 33000*3*4 = 509,600}
for evaluations.
To show the stability of $\metric$, we report the mean performance across the three seeds of oracle training.

\subsection{Results}
\label{sec:results}

\subsubsection{Concept Alignment.} Table~\ref{tab:main_results} shows the overall performance of the baselines in terms of $\metric$, where lower score indicates better performance.
First, we can observe that \ccd{} for \textit{concept alignment} is low for the original images; suggesting that the oracle is certain about its predictions.
Second, it can be inferred that Custom Diffusion performs poorly, while Textual Inversion (SD) outperforms the other methodologies except for the case of the learning styles. 
We attribute this behavior to differences in textual encoders. LDM trains the BERT-style textual encoder from scratch while SD uses pre-trained CLIP to condition the diffusion model.
CLIP contains vast image-text knowledge leading to better performance on learning objects but less flexibility to learn different styles as a concept.
Surprisingly, if we compare the \textit{concept alignment} performance with and without composite prompts, we observe that the performance further drops significantly for all baseline methodologies when composite prompts are used. 
This shows that existing concept learning methodologies find it difficult to maintain the concepts whenever the prompt contains the composition.

\subsubsection{Compositional Reasoning.} Previously, we discussed concept alignment on composite prompts. 
Table~\ref{tab:composition_results} summarizes the evaluations on composition tasks.
Here, we observe the complete opposite trend in results.
Custom Diffusion outperforms the other approaches across the composition categories.
This result shows the trade-off between learning concepts and at the same time maintaining compositionality in recent concept learning methodologies.
Moreover, CLIPScore estimates the better performance of the baselines compared to the original image-text pairs which are inaccurate. 

\subsubsection{Qualitative Results.} Figure~\ref{fig:qualitative_examples} provides the qualitative examples of the concept learning.
It can be inferred that Textual Inversion (LDM) learns the \concept{sketch} concept very well (the first row), while DreamBooth and Custom Diffusion struggle to learn it.
All baselines perform comparatively well in reproducing the learned concept (the second row).
Interestingly, in the case of compositions, DreamBooth and Custom Diffusion perform well with the cost of losing the concept alignment (the last two rows). 
At the same time, textual inversion approaches cannot reproduce the compositions (like, \textit{``Two V\textsuperscript{*}''}) but they maintain concept alignment.
Overall, these qualitative examples align with our quantitative results and strengthen our evaluation framework.

\subsubsection{Human Evaluations.} We perform Human Evaluations using Amazon Mechanical Turk for both types of evaluations: 1) concept alignment -- to measure the alignment between generated images and ground truth reference images on Domain\textsubscript{PACS} and Object\textsubscript{ImageNet}, and 2) compositional reasoning -- to measure the image-text alignment.
For concept alignment, we ask human annotators to rate the likelihood of the target image the same as three reference images.
While for compositional reasoning we simply ask the annotators to rate the likelihood alignment of the image and the corresponding caption. 
Table~\ref{tab:human_eval} summarizes the performance of prior and proposed ($\metric$) quantitative metrics \textit{w.r.t.} the Human Score. 
KID performs better for domains than objects as image dynamics varies a lot in domains.
\cite{kumari2022multi} proposed to use KID with LAION-retrieved concept images as a reference instead of ground truth due to the scarcity of reference images.
However, \conceptbed{} alleviates this limitation. Therefore, we use actual ground truth images to report KID which is more accurate.
It can be inferred that the \ccd{} is strongly correlated with human preferences and outperforms the prior evaluation metrics by a large amount.

\subsubsection{Percentage of highly aligned instances.} Using $\metric$, we can further measure the recall of the concept learning models.
DINO and KID metrics do not allow us to measure the recall. 
Hence, it becomes hard to investigate the actual quality of the generated images.
Table~\ref{tab:recall} shows the recall ($\frac{\mathrm{sample~with} \metric<=0.0}{\mathrm{total~samples}} * 100$) for the concept alignment shown in Table~\ref{tab:main_results}.
It can be inferred that Custom-Diffusion can work once in every four generation attempts.
While Textual Inversion will work at least once in every two attempts.
At the same time, when composition prompts are provided, Textual Inversion consistently maintains the concept alignment at the cost of achieving the composition alignment.

\begin{table}[]
    \centering
            \fontsize{9pt}{9pt}
    \selectfont
\setlength\tabcolsep{4pt}

    % \small
    % \resizebox{\linewidth}{!}{%
    \begin{tabular}{l | ccc | c}
        \toprule
        \textbf{Models} & \textbf{ConvNeXt} &  \textbf{Inception} & \textbf{ViT} & \textbf{Few-Shot} \\
        \midrule
        \textbf{TI (LDM)} & 0.0955 & 0.0773 & 0.1165 & 0.0823 \\ 
        \textbf{TI (SD)} & 0.0472 & 0.0201 & 0.0599 & 0.0489 \\ 
        \textbf{DB} & 0.0678 & 0.0485 & 0.0786 & 0.0596 \\ 
        \textbf{CD} & 0.2085 & 0.1845 & 0.2286 & 0.1384 \\ 
        \midrule
        \textbf{Correlation} & -0.9892 & -0.9888 & -0.9816 & -0.9763 \\
        \bottomrule

    \end{tabular}
    % }
    \caption{
Ablation.
Effect of different oracle models to measure concept alignment using $\metric$.
}
\label{tab:ablation}

\end{table}

\smallskip\noindent{\textbf{Generalization.}} 
Fine-tuned oracles cannot be generalized to unseen concepts; making \ccd{} unreliable on OOD concepts.
% Table~\ref{tab:ablation} shows that there is no impact on the choice of the classifier as an oracle. 
% However, fine-tuned oracles cannot be generalized to unseen concepts.
Hence, we propose to utilize a few-shot classifier (5-way 5-shot) instead, which can allow the generalization to unseen concepts while maintaining a high correlation (shown in Table~\ref{tab:ablation}). 
This shows the effectiveness of using confidence and \ccd{} as the alternative to the DINO, KID, and CLIP. 
\section{Related Work}

\textbf{Concept Learning.}
Concept learning encompasses various problem statements and approaches, depending on the perspective adopted. 
{Concept Bottleneck Models (CBMs)}~\cite{koh2020concept} and {Concept Embedding Models (CEMs)}~\cite{zarlenga2022concept} treat object attributes as concepts and propose classification strategies to identify these concepts. 
{Neuro Symbolic Concept Learner (NS-CL)}~\cite{mao2018the} aims to learn visual concepts by associating them with language semantics, enabling the model to perform visual question answering.
{Image Inversion Style Concept Learning} \cite{xia2022gan}, takes a different approach. 
Its objective is to invert a given concept image back into the latent space of a pre-trained model. 
However, text-based concept composition is not possible for such models.

\smallskip\noindent\textbf{Text-to-Image Generative Models.}
With advances in vector quantization~\cite{van2017neural} and diffusion modeling~\cite{rombach2022high}, text-to-image generation has improved its performance.
Notable works such as DALL-E~\cite{ramesh2021zero} train transformer models.
While current state-of-the-art, diffusion-based text-to-image models such as GLIDE~\cite{nichol2022glide}, LDM~\cite{rombach2022high}, and Imagen~\cite{saharia2022photorealistic}, have surpassed prior approaches (such as  StackGAN~\cite{zhang2017stackgan}, StackGAN++~\cite{zhang2018stackgan++}, TReCS~\cite{koh2021text}, and DALL-E~\cite{ramesh2021zero}) and achieved superior performance. 
Pixart-$\alpha$~\cite{chen2023pixart} and ECLIPSE~\cite{patel2023eclipse} further enhances T2I methods without depending on heavy compute.
Additionally, as shown by \cite{saxon-wang-2023-multilingual}, these T2I models also have multilingual concept understanding to a certain extent.

\smallskip\noindent\textbf{Text-to-Image Concept Learning.} 
Text-conditioned diffusion models, such as LDM, have demonstrated their potential for learning novel visual concepts with only a few reference images. 
Textual Inversion~\cite{gal2022image} proposes learning the embedding corresponding to the placeholder (V\textsuperscript{*}) through optimization. 
DreamBooth~\cite{ruiz2022dreambooth} suggests optimizing the UNet parameters instead of optimizing the placeholder embedding. 
Custom Diffusion~\cite{kumari2022multi} combines both approaches by optimizing the placeholder and key/value weights from the cross-attention layers for faster concept learning. 
These concept learners are essentially text-conditioned diffusion models and inherit the same limitations of diffusion models. 
One limitation is the overfitting of concepts and language drift. 
By optimizing model parameters on a handful of reference images, it is highly likely that the model might overfit the given concept and cannot maintain compositionality.
Therefore, in this paper, we propose \conceptbed{} for systematic evaluations.

\smallskip\noindent\textbf{Text-to-Image Generative Model Evaluations.}
Evaluating generative models is not widely studied. 
The FID \cite{heusel2017gans} score is commonly used to measure generated image quality. 
CLIPScore \cite{hessel2021clipscore} is another popular evaluation metric for reference-free image-text alignment. 
Another study focuses on compositional evaluations of text-to-image models on small subsets (CU-Birds and Oxford-Flowers) \cite{park2021benchmark}. 
DALL-Eval \cite{cho2022dall} evaluates reasoning skills on synthetic datasets and social biases of text-to-image generative models.
DALL-Eval, VISOR \cite{gokhale2022benchmarking}, LAYOUTBENCH~\cite{cho2023diagnostic} evaluates spatial reasoning abilities.
Parallel work T2I CompBench~\cite{huang2023t2i} also adopts the idea of VQA for accurate composition evaluations.
Although text-to-image model evaluations are well-explored, they lack concept-specific assessments and cannot be used for evaluating concept learning. 
Therefore, \conceptbed{} attempts to overcome this gap in evaluations of novel visual concept learning abilities.
\section{Conclusion}
In this paper, we introduce a novel benchmark called \conceptbed{} designed to assess the efficacy of text-conditioned diffusion models in learning new concepts (\textit{a.k.a.} personalized T2I). 
The \conceptbed{} benchmark encompasses an end-to-end evaluation pipeline, a comprehensive concept library, and a novel \underline{C}oncept \underline{C}onfidence \underline{D}eviation (\ccd{}) evaluation metric.
We conduct evaluations based on two key criteria: concept alignment and composition alignment. 
Through extensive experiments, we demonstrate that existing text-conditioned diffusion model-based concept learners exhibit significant limitations in their performance.
We perform human evaluations to validate the effectiveness of our proposed evaluation metric ($\metric$), which showcases a strong correlation with human preferences. 
This finding positions $\metric$ as a viable alternative to human judgments, enabling large-scale and comprehensive evaluations.
\conceptbed{} represents the first large-scale concept-learning dataset that facilitates precise and accurate evaluations of personalized text-to-image generative models.

\section*{Acknowledgments}
This work was supported by NSF RI grants \#1750082 and \#2132724, and a grant from Meta AI Learning Alliance. The views and opinions of the authors expressed herein do not necessarily state or reflect those of the funding agencies and employers.

% \clearpage
\bibliography{aaai24}

\clearpage
\appendix
\section{Social Impact}

In this paper, we introduce \conceptbed{}, a novel benchmark and evaluation framework designed for conducting comprehensive studies on few-shot Concept Learning using T2I diffusion models. 
Previous evaluations of recent works in this field have been limited to a small number of test concepts, thus hindering our understanding of their practical applicability. 
Through our benchmark, we demonstrate that while current concept learners exhibit impressive performance, a substantial gap remains that must be addressed.
As pioneers in constructing this extensive evaluation set, we anticipate that future research will incorporate a broader range of potential concepts. 
Additionally, we propose a novel evaluation metric and framework that can be applied to any concept learning setting, extending its efficacy beyond the confines of \conceptbed{} dataset.
Ultimately, this research directly contributes to the advancement of Human-Level Artificial Intelligence (HLAI) objectives, fostering the development of more robust and capable systems.

% \section{\conceptbed{} Resources}
% \input{tables/appendix/accecibility}
% We release the code (a collection of concept learners), data, and results (randomly selected from over 200,000 images) for readers to refer to or browse through. 

\section{Extended Related Work}

\textbf{Evaluations of T2I Concept Learners.} Previous studies on concept learning have conducted evaluations and model comparisons using their own test sets. 
For instance, Textual Inversion~\cite{gal2022image} employed approximately 20 concepts with around 27 unique compositions, while DreamBooth~\cite{ruiz2022dreambooth} utilized 30 concepts with 50 unique compositions. Custom Diffusion~\cite{kumari2022multi}, on the other hand, employed 10 concepts with 24 unique compositions. 
Notably, these works were evaluated on a relatively small subset of concepts and a limited list of compositions. 
In order to address the limitations associated with a centralized evaluation set, we introduce the \conceptbed{} dataset, which consists of 284 concepts and over 33000 compositions. 
Additionally, we present an automated procedure for concept and composition collection, enabling the creation of large-scale datasets.

\textbf{Downstream Applications of Diffusion Models.} In addition to concept learning, diffusion models have demonstrated potential for various downstream applications. 
For example, approaches such as prompt-to-prompt~\cite{hertz2022prompt} and DiffEdit~\cite{couairon2022diffedit} have been proposed for image editing tasks. 
In another case, diffusion-generated images have shown improvements in ImageNet accuracy~\cite{azizi2023synthetic}. 
Furthermore, methods similar to textual inversion have been found to enhance few-shot classification performance~\cite{trabucco2023effective}.

\textbf{Out-of-Distribution Detection and Domain Adaptation/Generalization.} While the research directions of out-of-distribution detection and domain adaptation/generalization have been explored independently to a significant extent, they share a common focus on measuring and controlling model confidence. 
Prior works have employed various confidence quantification methods, including: 1) Expected Calibration Error (ECE), which is a popular metric for assessing classifier calibration by measuring the difference between model accuracy and its probability~\cite{naeini2015obtaining}, and 2) Expected Uncertainty Calibration Error (UCE), a recently proposed metric that quantifies the miscalibration of uncertainty by calculating the difference between model error and its uncertainty~\cite{laveswell}. 
Given the high variance observed in diffusion models with respect to hyperparameters, we introduce a novel method, leveraging the \conceptbed{} dataset, to quantify generation variances and measure deviations using $\metric$. 
ECE and UCE can serve as alternative metrics for quantifying deviations and evaluating concept learners.
Our experimental results in Appendix~\ref{sec:all_metrics} demonstrate that ECE performs equally well as $\metric$ in assessing concept alignment. 
In the context of concept alignment, ECE and UCE can be computed based on generated concept-specific images, without considering the performance on the ground truth target images. 
Lower values of these metrics indicate better performance, albeit at the cost of explainability regarding the source of errors (e.g., overconfidence or lack of confidence in the model). 
To address these potential ambiguities, we propose $\metric$, which measures the discrepancy in probabilities between ground truth and generated concept-specific images, thereby facilitating a more nuanced understanding of the limitations of concept learners.

\section{Preliminaries on text-conditioned diffusion models}
\textbf{Diffusion Models:} The training procedure of Stable Diffusion can be described as follows: 
given a training pair $(\mathcal{I}, y)$, the input image $\mathcal{I}$ is first mapped to a latent vector $z$ and get a variably-noised vector $z^t:= \alpha^t z^{t-1} + \sigma^t \epsilon $, 
where $\epsilon \sim \mathcal{N}(0,1)$ is a noise term and $\alpha^t, \sigma^t$ are terms that control the noise schedule and sample quality. 
At training time, the time-conditioned UNet is optimized to predict the noise $\epsilon$ and recover the initial $z$, via conditioning on the text prompt $y$, the model is trained with a squared error loss on the predicted noise term as follows:
\begin{align}
    \mathcal{L}_{\text{diffusion}} = \mathbb{E}_{z, \epsilon \sim \mathcal{N}(0,1),t,y} \Big[||\epsilon - \epsilon_{\theta}(z^{t},t,y)||_{2}^{2} \Big]
\end{align}
where $t$ is uniformly sampled from $\{1,\dots,T\}$. 

At inference time, Stable Diffusion is sampled by iteratively denoising $z^T \sim \mathcal{N}(0,I)$ conditioned on the text prompt $y$. 
Specifically, at each denoising step $t = 1,\dots,T$, $z^{t-1}$ is obtained from both $z^t$ and the predicted noise term of UNet whose input is $z^t$ and text prompt $y$. 
After the final denoising step, $z^0$ will be mapped back to yield the generated image $\mathcal{I}$.

\textbf{Textual-Inversion (TI):} TI uses the pre-trained Stable Diffusion and fine-tunes it to learn the specific concepts using a few images. 
Given a small set of images depicting the target concept $\mathcal{X}_c = \{x_c^i; i \in \{0, ..., m\} \}$, and with the rare-token $y_k$ (i.e., V\textsuperscript{*}), we want to learn the embedding corresponding to $y_k$. 
This input-conditioned text can be represented as ``A photo of a V\textsuperscript{*}''. 

TI follows the exact same process of Stable Diffusion. 
Unlike Stable Diffusion, TI optimizes the text conditional encoder ($C_\phi$) with respect to the rarely occurring token $y_k$ using the Latent Diffusion Model (LDM) objective function:

\begin{align*}
    \mathcal{L}_{\text{TI}} = \mathbb{E}_{z, \epsilon \sim \mathcal{N}(0,1),t,y} \Big[||\epsilon - \epsilon_{\theta}(z^{t},t,C_\phi(y))||_{2}^{2} \Big]
\end{align*}

Note that $z^t$ is the noised $x$ where $x \in X_c$.
Intuitively, the objective is to correctly remove the added noise (while training) and optimize $C_\phi$ with respect to $y_k$.
At inference time, a random noise tensor is sampled and a text prompt (containing the rare-token $y_k$) is used to generate the image it using fine-tuned $C_\phi$.

\textbf{DreamBooth:} While textual-inversion can be used to learn various concepts depending on the training images and corresponding set of text prompts, DreamBooth is proposed to learn the specific properties of the target subject: ``A photo of a V\textsc{*} dog''.
In the case of DreamBooth, we do not optimize $C_\phi$ and instead, it optimizes $\epsilon_\theta$.

To overcome the challenges (overfitting and language drift) of fine-tuning the full model, DreamBooth contains the class-specific prior-preserving loss. 
Essentially, this method uses the pre-trained diffusion model generated samples ($X_{pr} = \{\hat{x_i}; \hat{x_i}=f(\epsilon,c_{pr})\}$) to supervise the training.
Here, $\epsilon \sim \mathcal{N}(0,1)$ and conditioning vector $c_{pr} = C_\phi(``<\mathrm{concept-name}>'')$.
Therefore, the proposed loss becomes:

\begin{align*}
    & \mathcal{L}_{\text{DB}} = \mathbb{E}_{z, \epsilon \sim \mathcal{N}(0,1),t,y,x \in X_c}   \Big[||\epsilon - \epsilon_{\theta}(z^{t},t,C_\phi(y))||_{2}^{2} \Big] \\& + \lambda* \mathbb{E}_{z, \epsilon \sim \mathcal{N}(0,1),t,y_{pr}, \hat{x} \in X_{pr}}  \Big[||\epsilon - \epsilon_{\theta}(z^{t},t,C_\phi(y_{pr}))||_{2}^{2} \Big]
\end{align*}

\textbf{Custom-Diffusion:}
For single-concept learning, Custom Diffusion is essentially the combination of Textual Inversion and DreamBooth.
The objective function of Custom Diffusion is the same as DreamBooth but instead of optimizing whole UNet (i.e., $\epsilon_\theta$), Custom Diffusion optimizes the embedding corresponding to V\textsuperscript{*} from $C_\phi$ and key/value weights from Cross Attention Layers of the UNet model.

\section{\conceptbed{} Dataset}

\subsection{ImageNet Dataset Generation Pipeline}
As mentioned in the main text, ImageNet contains 1000 classes but not all of them are used in day-to-day interactions.
Moreover, performing experiments on each of these 1000 classes is computationally very extensive as one needs to train 4000 models and generate 400,000 images. 
Therefore, it is important to filter out highly used concepts in daily life.
To measure the real-life- importance we check if any concept (such as \concept{dog}) is the subject of the caption prompt in the whole visual genome dataset. 
If there exist at least 10 captions having the concept as subject then we add the concept in \conceptbed{} library.
Additionally, the concept learning methodologies can learn new concepts using as little as 4 images.
Using all ImageNet images as training data can potentially add more noise as these images are not high resolution.
Hence, we further filter out the top 100 images based on the percentage of the object pixels (with a ratio of at least 0.4) within the image. 
We provide the Algorithm~\ref{alg:data_algo} for readers' understanding of the data generation pipeline. 
It is worth noting that, this pipeline can be used to extend the \conceptbed{} and even to train the future concept learning methodologies.

\begin{algorithm}[t]
    \caption{\conceptbed{} Object-Concepts Collection Pipeline}
    \label{alg:data_algo}
    \small
    \begin{algorithmic}[0]
        % \REQUIRE $n \geq 0 \vee x \neq 0$
        \STATE \textbf{Input:} $Y_{VG} = \mathrm{Visual~Genome~Captions}$;\\ 
        \STATE \quad \quad $C_{ImageNet} = \{(c, X_c)\}_1^{N}$;  \\
        \STATE \textbf{Output:} Estimated $C_{\conceptbed{}} = \{(\hat{X}_c, c)\}_1^{M}$
    \end{algorithmic}
    \begin{algorithmic}[1]
        \STATE \textbf{Initialize:} $C_{ConceptBed} = []$; $M = 0$
        \FOR{$(c, X_c) \in C_{ImageNet}$}
            \STATE $count = 0$
            \FOR{$y \in Y_{VG}$}
                \IF{$subject(y) == c$}
                    \STATE count = count + 1
                \ENDIF
            \ENDFOR
            \IF{count>=10}
                \STATE $\hat{X}_c = []$
                \FOR{$x \in X_c$} 
                % \COMMENT{// Optimization is ignored for simplicity of the pseudo-code}
                    \STATE $area = \frac{\#~of~pixels(c)}{\# of~total~pixels}$
                    \IF{$area>=0.4$}
                        \STATE $\hat{X}_c \gets x$
                    \ENDIF
                \ENDFOR
                \STATE $\hat{X}_c = sorted(\hat{X_c})$
                \STATE $C_{\conceptbed{}} \gets (\hat{X}_c[:100], c) $
                \STATE $M = M + 1$
            \ENDIF
        \ENDFOR
    \end{algorithmic}
    
\end{algorithm}

\subsection{Concept Statistics}
Our dataset, \conceptbed{}, comprises a total of 284 concepts.
Among these concepts, 200 are sourced from the CUB dataset, 80 from ImageNet, and 4 from PACS. 
The concepts and their respective categories are presented in Table~\ref{tab:concepts}. 
We use the CUB dataset for attribute-level analysis, while the ImageNet concepts are included to ensure a diversity of concepts.

\subsection{Composition Categorization}

\begin{figure}
    \begin{subfigure}{.5\textwidth}
      \centering
      \includegraphics[width=\linewidth]{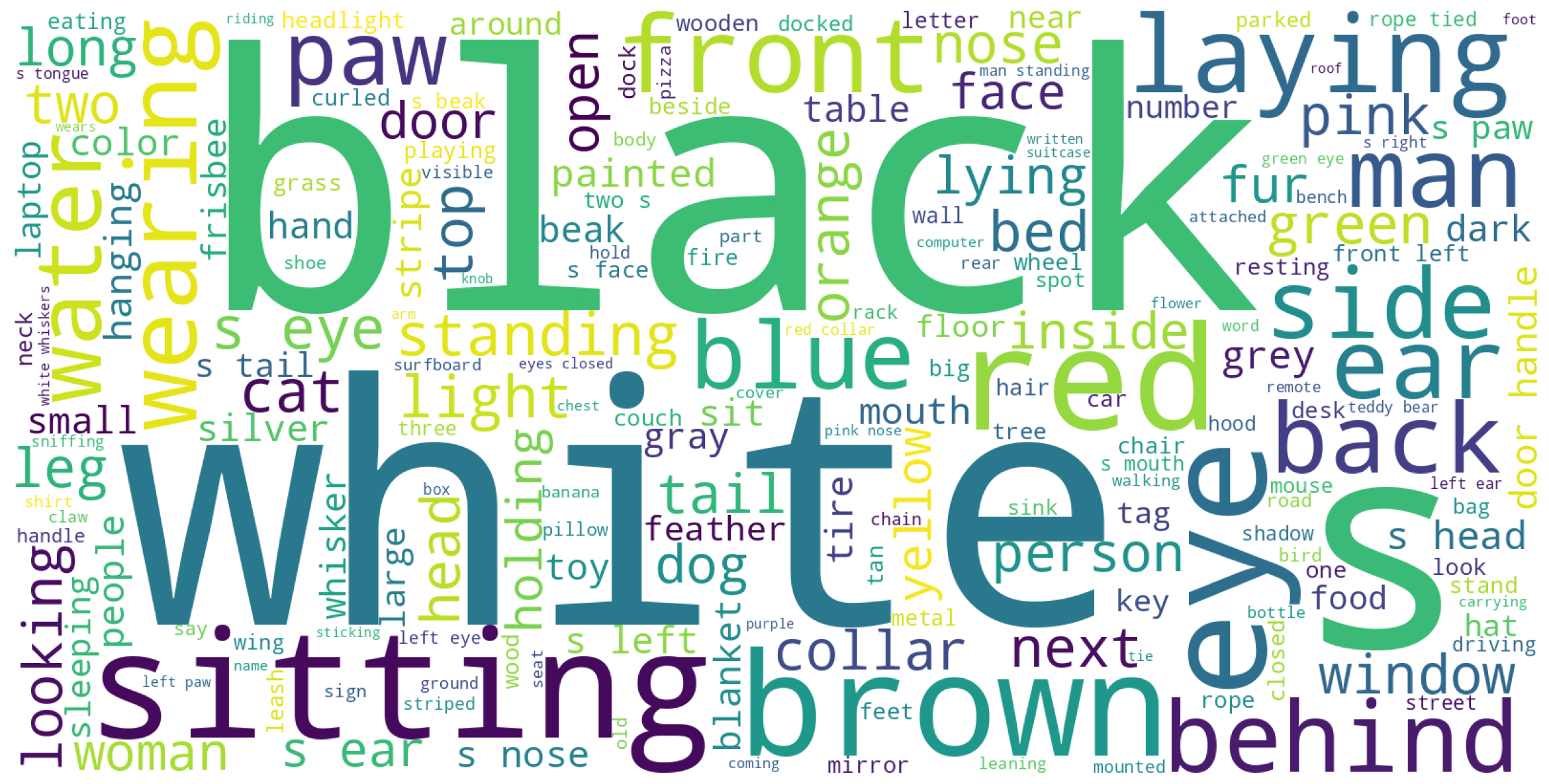}
      \label{fig:wordcloud}
    \end{subfigure}
    \begin{subfigure}{.5\textwidth}
      \centering
      \includegraphics[width=\linewidth]{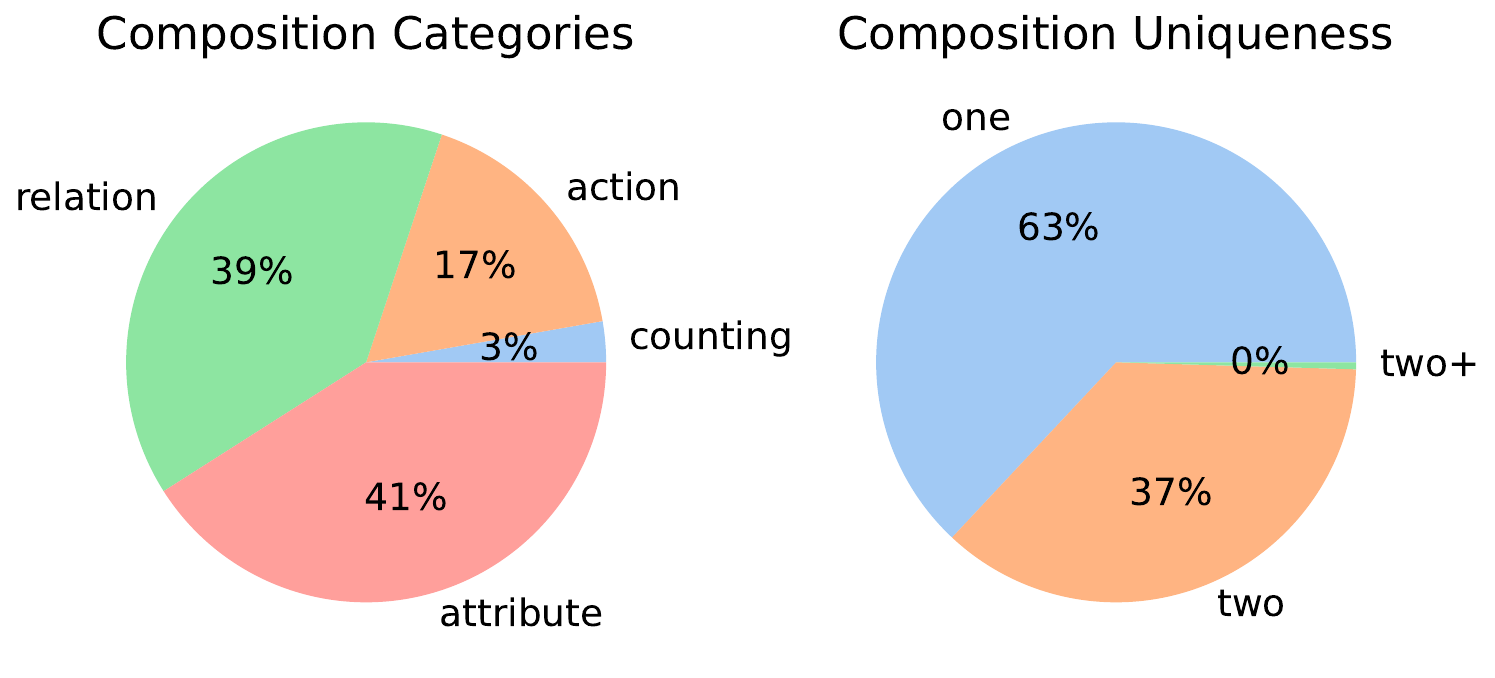}
      \label{fig:all_comp_stat}
    \end{subfigure}
    \caption{Top-most figure shows the world-cloud of the \conceptbed{} compositions. Bottom-left figure shows the statistics of composition categories. 
    Bottom-right figure shows the statistics of multiple composition categories combined together to generate composite text prompts.}
    \label{fig:stats}
\end{figure}

\begin{figure}
    \centering
    \includegraphics[width=\linewidth]{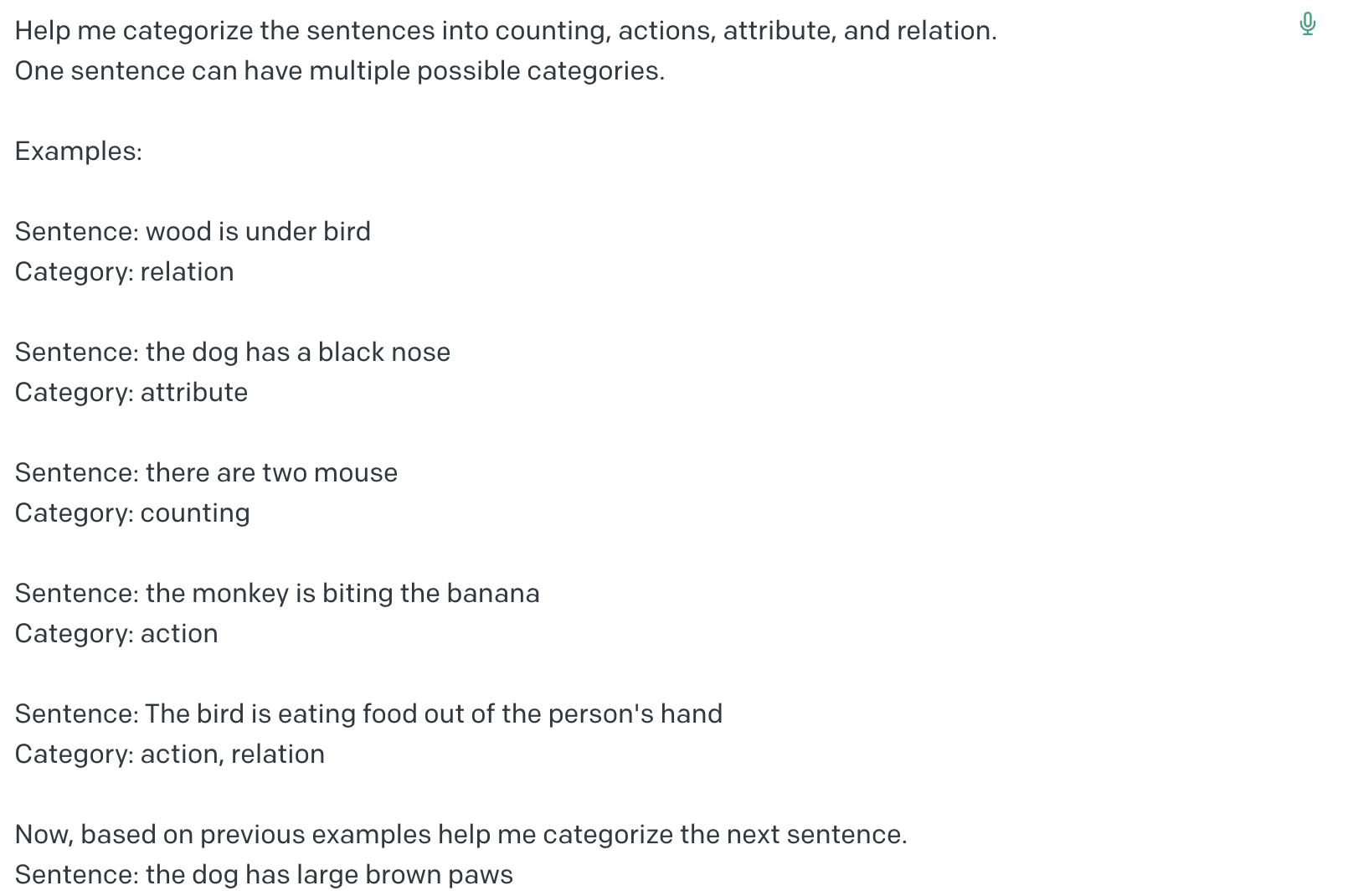}
    \caption{Screenshot depicting few-shot classification using GPT3. We keep the instruction and few-shot examples constant and change the target phrase to get the corresponding categories.}
    \label{fig:gpt3_composition}
\end{figure}

We leverage the Visual Genome dataset to create composite prompts for each of the 80 ImageNet concepts. 
This process yields over 33,000 compositions, resulting in a rich variety of prompts.
Table~\ref{tab:concept_stats} provides detailed statistics on the compositions for each concept. 
Furthermore, Figure~\ref{fig:stats} illustrates the distribution of composition categories within the \conceptbed{} dataset. 
For the sake of simplicity, \conceptbed{} contains composite prompts that combine up to two different compositions. 
To determine the composition type, we employ the GPT3 (text-davinci-003) model for few-shot classification. 
Figure~\ref{fig:gpt3_composition} showcases the instruction and in-context examples used to categorize each text phrase.

\subsection{Question Generations}
\begin{table}[ht]
    \centering
    \small
    \resizebox{\linewidth}{!}{%
    
    \begin{tabular}{@{}l l@{}}
        \toprule
        \textbf{Caption} & \textbf{Generated Questions} \\
        \midrule
        \multirow{2}{*}{two \concept{bird}s standing on branches} & are there two birds ? \\
        & are there branches ? \\
        \midrule
        \multirow{2}{*}{tongue hanging out a \concept{dog}'s mouth} & is there tongue ? \\
        & is there a dog's mouth ?\\
        \midrule
        \multirow{2}{*}{\concept{cat} is licking herself} & is there cat ? \\
        & is the cat licking herself ? \\
        \midrule
        \multirow{2}{*}{the \concept{monkey} is holding onto a red bar} & is there the monkey ? \\
        & is there a red bar ? \\
        \midrule
        \multirow{2}{*}{Propeller blade on an \concept{aircraft}} & is there propeller blade ? \\
        & is there an aircraft ? \\
        \midrule
        \multirow{2}{*}{the vine is around \concept{clock}} & is there the vine ? \\
        & is there clock ? \\
        \midrule
        \multirow{2}{*}{man driving the \concept{boat}} & is there man ? \\
        & is there the boat ? \\
        \bottomrule
        
    \end{tabular}%
    }
    \caption{\textbf{Question Generation.} Examples of generated existence-related questions from captions.}
    \label{tab:vqa}
\end{table}
Rather than relying solely on image-text similarity, we evaluate compositions through VQA performance using synthetically generated boolean questions (i.e., \textit{yes} or \textit{no}) based on the composite text phrases. 
To create these questions, we manually filter out salient words such as nouns, attributes, and verbs, and formulate questions corresponding to each of these words. 
This process enables the creation of existence-related questions where the ground truth answer is always \textit{yes}. 
Table~\ref{tab:vqa} provides examples of the questions generated for different composite text prompts.

\section{Experimental Setup}
\begin{table*}[ht]
    \centering
    \resizebox{\textwidth}{!}{%
    
    \begin{tabular}{@{}l | cccc@{}}

        \toprule
        \textbf{Hyper-Parameter} & \textbf{Textual Inversion (LDM)} & \textbf{Textual Inversion (SD)} & \textbf{DreamBooth} & \textbf{Custom Diffusion} \\
        \midrule
         Base Model & LDM & Stable Diffusion - v1.5 & Stable Diffusion - v1.5 & Stable Diffusion - v1.5 \\
         Optimized & V\textsuperscript{*} & V\textsuperscript{*} & UNet & V\textsuperscript{*} + CrossAtten(k,v) \\
         Optimization Steps & 3000 & 3000 & 400 & 250\\
         Learning Rate & 5e-4 & 5e-4 & 5e-6 & 1e-5 \\
         Place-Holder Token & * & $<$object$>$ & sks & $<$new1$>$ \\
         Regularizer & - & - & \greencheck & \greencheck \\
         Regularization Images & - & - & 200 & 200 \\
         \midrule
         \# if inference steps & 50 & 50 & 50 & 50 \\
         Guidance Scale & 7.5 & 7.5 & 7.5 & 7.5 \\
         Noise Scheduler & - & PNDMScheduler & PNDMScheduler & PNDMScheduler \\
        \bottomrule
        
    \end{tabular}%
    }
    \caption{\textbf{Hyper-parameters.} The table summarizes the different hyper-parameter settings for all baselines considered in this work to help the reproducibility of results. }
    \label{tab:baseline_hyper}
\end{table*}
\begin{table*}[ht]
    \centering
    % \resizebox{\textwidth}{!}{%
    
    \begin{tabular}{@{}l | cccc@{}}
        \toprule
        \textbf{Hyper-parameters} & \textbf{Domain\textsubscript{PACS}} & \textbf{Object\textsubscript{ImageNet}} & \textbf{Fine-Grained\textsubscript{CUB}} & \textbf{Compositions} \\
        \midrule
        Model Architecture & ResNet18 & ConceNeXt-base & - & ViLT \\
        % Pre-trained Weights & \href{https://github.com/huggingface/pytorch-image-models/blob/main/timm/models/resnet.py}{timm} & \href{https://github.com/huggingface/pytorch-image-models/blob/main/timm/models/convnext.py}{timm} & - & \href{https://huggingface.co/dandelin/vilt-b32-finetuned-vqa}{dandelin/vilt-b32-finetuned-vqa} \\
        Pre-training Dataset & ImageNet & ImageNet & - & MSCOCO, GCC, SBU, VG \\
        \# of target concepts & 4 & 80 & 200/112 & 1 \\
        Objective Function & NLL & NLL (w outlier exposure) & NLL & NLL \\
        \bottomrule
        
    \end{tabular}%
    % }
    \caption{\textbf{Oracle Hyper-parameters.} This table summarizes the different hyper-parameters used for Oracles. We first take the pre-trained model weights and then fine-tune them on target concepts from the \conceptbed{} dataset. Here, NLL refers to the Negative Log-Likelihood.}
    \label{tab:orcale_hyper}
\end{table*}

Table~\ref{tab:baseline_hyper} presents the hyperparameter details for the baseline methodologies employed in our benchmarking process. 
To ensure fair comparisons, we generate images for all concepts using the same number of inference steps and guidance scale. 
Specifically, Textual Inversion (SD), DreamBooth, and Custom Diffusion utilize the Stable Diffusion V1.5 pre-trained model\footnote{\url{https://huggingface.co/runwayml/stable-diffusion-v1-5}}.
Regarding Textual Inversion, we explore two variants: 1) Latent Diffusion Model-based, and 2) Stable Diffusion-based. 
By incorporating different pre-trained models, we aim to investigate their impact on learning novel concepts, and our findings reveal significant differences. 
For instance, Textual Inversion (LDM) outperforms Textual Inversion (SD) when learning style as a concept, while the SD version excels in adapting object-level concepts.

Table~\ref{tab:orcale_hyper} outlines the hyperparameter settings for each oracle. 
It is important to note that we can employ any type of classifier as an oracle, as elaborated in the subsequent section. 
In our approach, we initially take pre-trained models and further, fine-tune them on \concept{} concepts using the negative log-likelihood objective. 
However, it is well-established that classifiers may exhibit misclassification tendencies with high confidence. 
To address this, for Objects\textsubscript{ImageNet}, we additionally incorporate an outlier-exposure objective function~\cite{hendrycksdeep}.

\section{Human Annotations}

\begin{figure*}
    \centering
    \includegraphics[width=\textwidth]{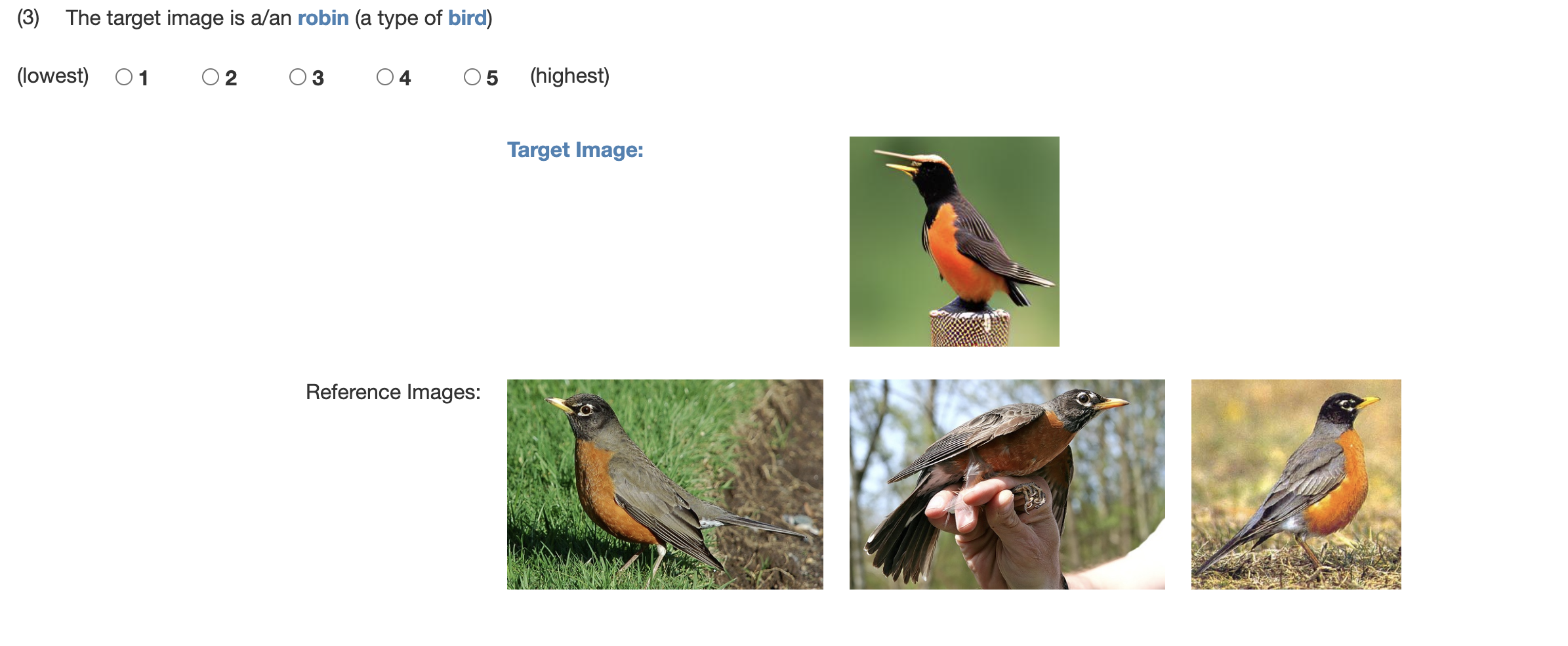}
    \caption{An example of human annotation for determining the concept alignment for \textbf{object-specific concepts}.}
    \label{fig:object_hit}
\end{figure*}

\begin{figure*}
    \centering
    \includegraphics[width=\textwidth]{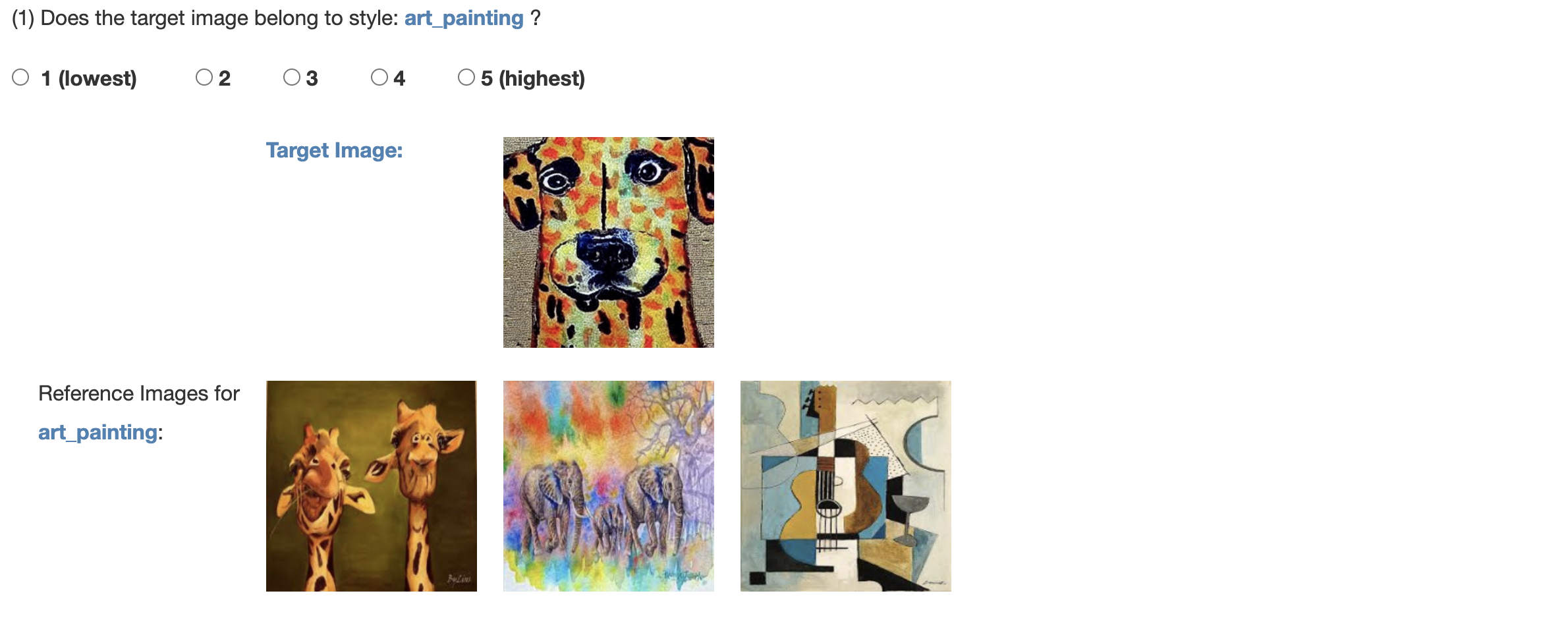}
    \caption{An example of human annotation for determining the concept alignment for \textbf{ style-specific concepts}.}
    \label{fig:style_hit}
\end{figure*}

\begin{figure*}
    \centering
    \includegraphics[width=\textwidth]{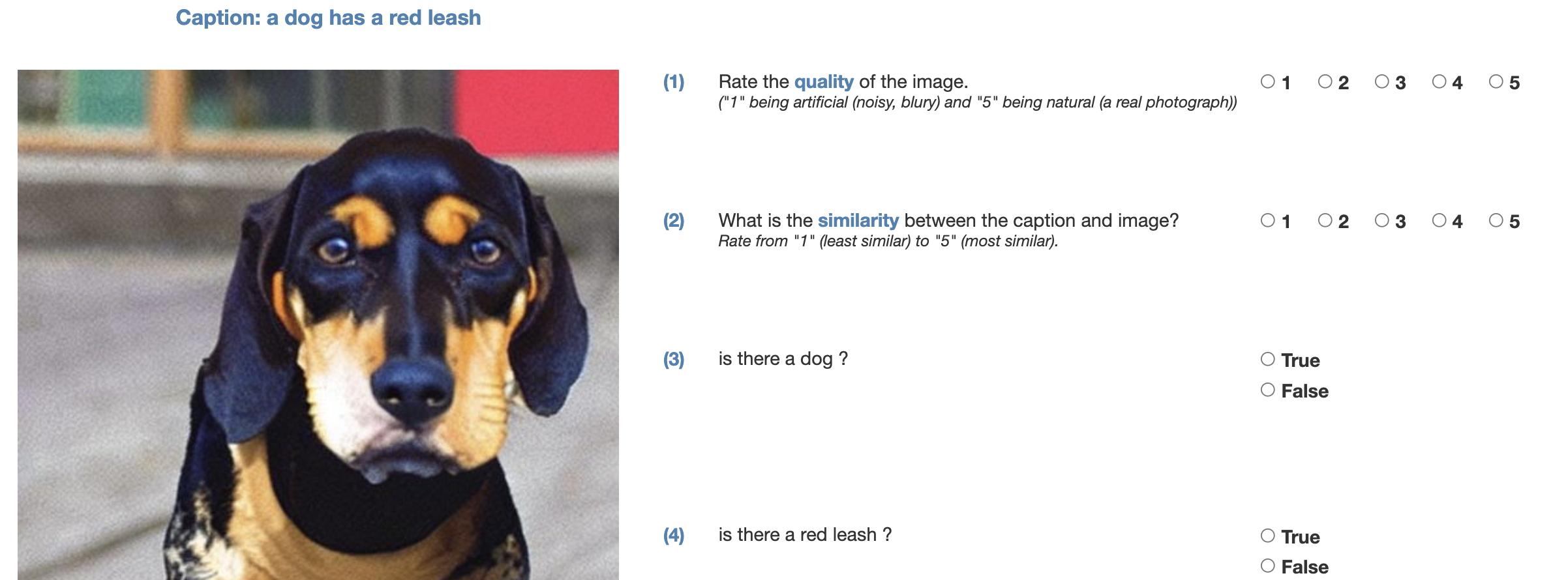}
    \caption{The example of Human Annotation for determining the \textbf{image-text alignment}.}
    \label{fig:composition_hit}
\end{figure*}

To evaluate the effectiveness of our \conceptbed{} evaluation framework, we conducted human evaluations consisting of three distinct studies: object-based concept similarity, style-based concept similarity, and traditional image-text similarity for evaluating compositions.
For the object and style-based concept similarity, we asked participants to rate the likelihood of the target image being the same as three reference images on a scale of 1 to 5. 
A rating of 1 indicated the least similarity, while a rating of 5 indicated an exact match in terms of concept. 
We ensured that human annotators did not compare generated images from different concept learning strategies; instead, they rated each image independently.
Regarding the composition evaluation, we simply asked annotators to rate the image-text similarities on the same 1-5 scale, with 1 representing the least similarity and 5 representing an exact match.
Figures~\ref{fig:object_hit},~\ref{fig:style_hit}, and~\ref{fig:composition_hit} present screenshots of the MTurk interface used for each type of human evaluation.

To ensure comprehensive coverage, we randomly selected 100 generated images and obtained evaluations from three unique workers for each image. 
This resulted in a total of 900 evaluations from human annotators. 
To assess the relationship between human evaluations and various baseline evaluation metrics, as well as our $\metric$ method, we computed Pearson's correlation. 
Our findings indicate \textbf{a strong correlation between the human evaluations and our \ccd{} evaluation metric}.

\section{Ablations}
\subsection{Different Confidence Measures}
\label{sec:all_metrics}

Table~\ref{tab:all_metrics} presents a comprehensive comparison of various confidence quantification metrics employed in Out-Of-Distribution (OOD) detection. 
Notably, all these metrics outperform the baseline metrics DINO and KID, as evidenced by their consistently high correlation scores, reaching an absolute high correlation of at least $0.9$.
This implies that our evaluation framework supports multiple metrics to measure the alignment as we are performing supervised learning to train the oracles.
Importantly, Accuracy and ECE measure the performance of a large collection of generated images.
While MSP and $\metric$ measure the performance at the instance level, which is more useful in practical scenarios where we don't have access to a lot of generated images to estimate the performance.
Although MSP also achieves a high correlation, in some cases, there might be a chance that an oracle can predict the wrong class with high confidence (as it is class-label independent).  
For instance, MSP on domain alignment leads to only a $0.14$ correlation with human preferences.
Hence, conditional probability is important to measure the instance-level alignment.
It is worth noting that the negative sign in the correlation coefficients stems from the inherent differences between the nature of these metrics. 
Specifically, lower values of $\metric$, and ECE indicate better performance, while higher scores in human evaluations indicate superior performance.

\subsection{Choice of Classifiers for Oracles}

In Table~\ref{tab:diff_cls}, we explore the impact of utilizing different types of classifiers as oracles. 
Our analysis encompasses four distinct classifiers, each characterized by an increasing number of parameters. 
Intriguingly, the choice of classifier appears to have a negligible effect, as $\metric$ consistently demonstrates strong correlations with human scores, surpassing a Pearson's correlation of at least $-0.98$.

\begin{table}[!h]
    \centering
    \small
    \resizebox{\linewidth}{!}{%
    
    \begin{tabular}{@{}l cccc@{}}
        \toprule
        \textbf{Models} &   \textbf{Accuracy ($\uparrow$)} &       \textbf{MSP ($\uparrow$)} &       \textbf{ECE ($\downarrow$)} &     \textbf{$\metric (\downarrow)$} \\ %& \textbf{Human Score} \\
        \midrule
        Textual Inversion (LDM)              &     80.07\% &  0.8734 &  0.0755 &     0.0955 \\
Textual Inversion (SD)    &     84.30\% &  0.9022 &  0.0623 &     0.0472 \\
DreamBooth &     83.17\% &  0.8923 &  0.0647 &     0.0678 \\
Custom Diffusion      &     69.73\% &  0.8311 &  0.1382 &     0.2085 \\
\midrule
Original                          &     89.31\% &  0.9276 &  0.0436 &    -0.0000 \\

        \bottomrule
        
    \end{tabular}%
    }
    \caption{\textbf{Possible Confidence Quantification Metrics.} This table summarizes the results using different existing confidence quantification metrics on \conceptbed{} dataset on 80 concepts from ImageNet. Here, MSP refers to the Maximum Softmax Probability and ECE refers to Expected Calibration Error.}
    \label{tab:all_metrics}
\end{table}

\section{Qualitative Results}
\begin{figure*}
    \includegraphics[width=\textwidth]{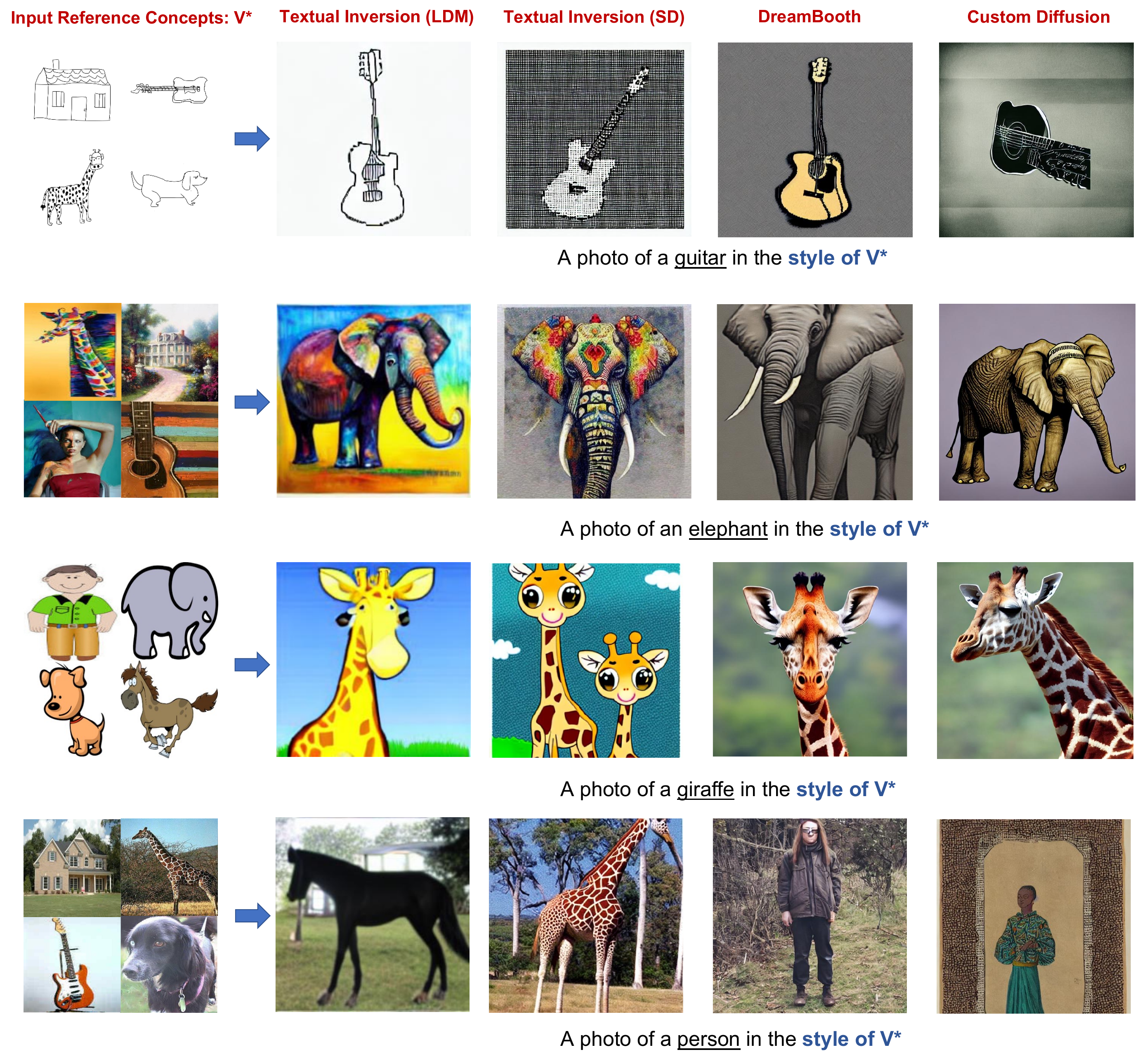}
    \caption{Qualitative examples of the style-specific four concepts.}
    \label{fig:qual_style}
\end{figure*}

\begin{figure*}
    \includegraphics[width=\textwidth]{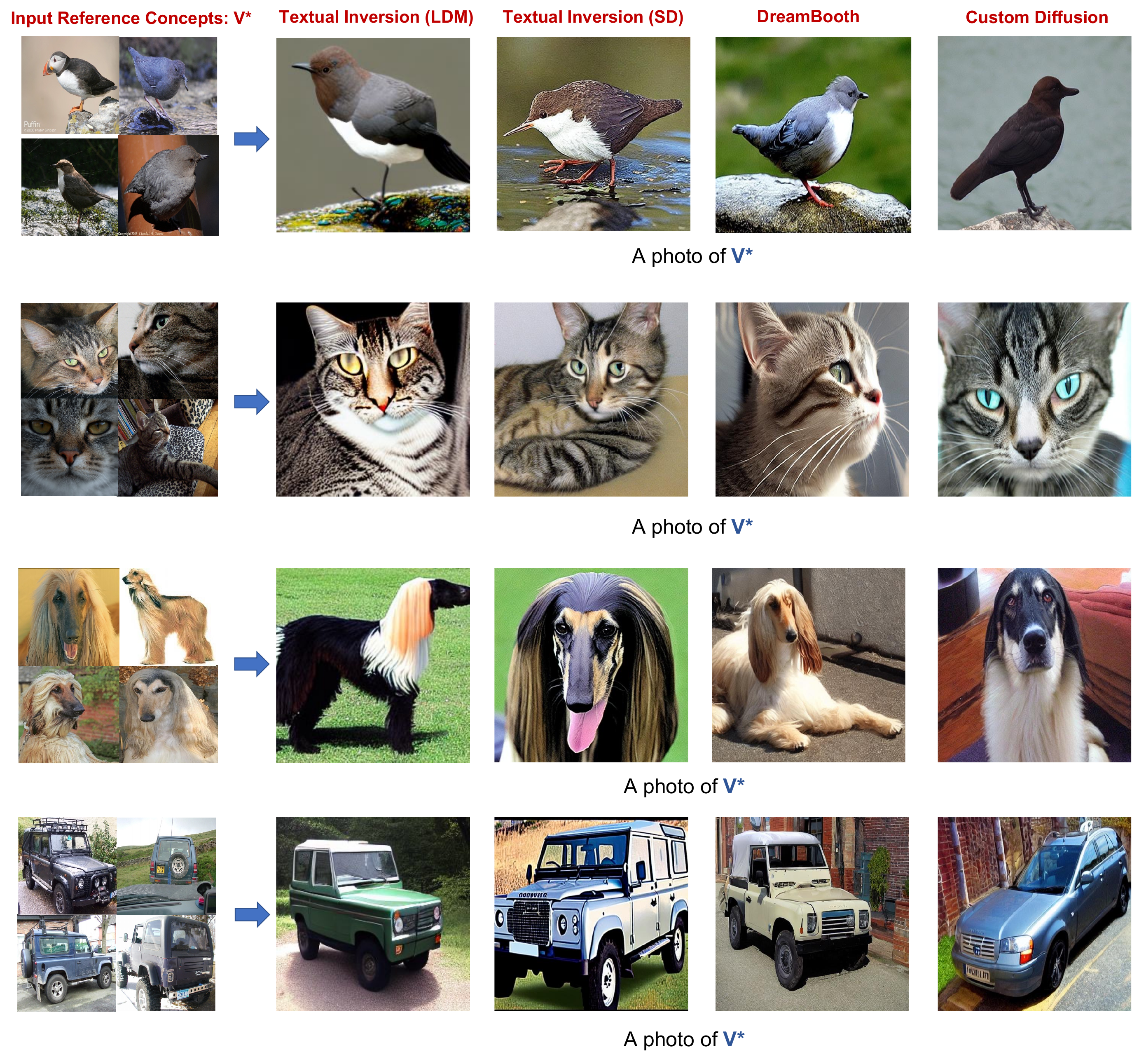}
    \caption{Qualitative examples of the object-specific four concepts.}
    \label{fig:qual_object}
\end{figure*}

\begin{figure*}
    \includegraphics[width=\textwidth]{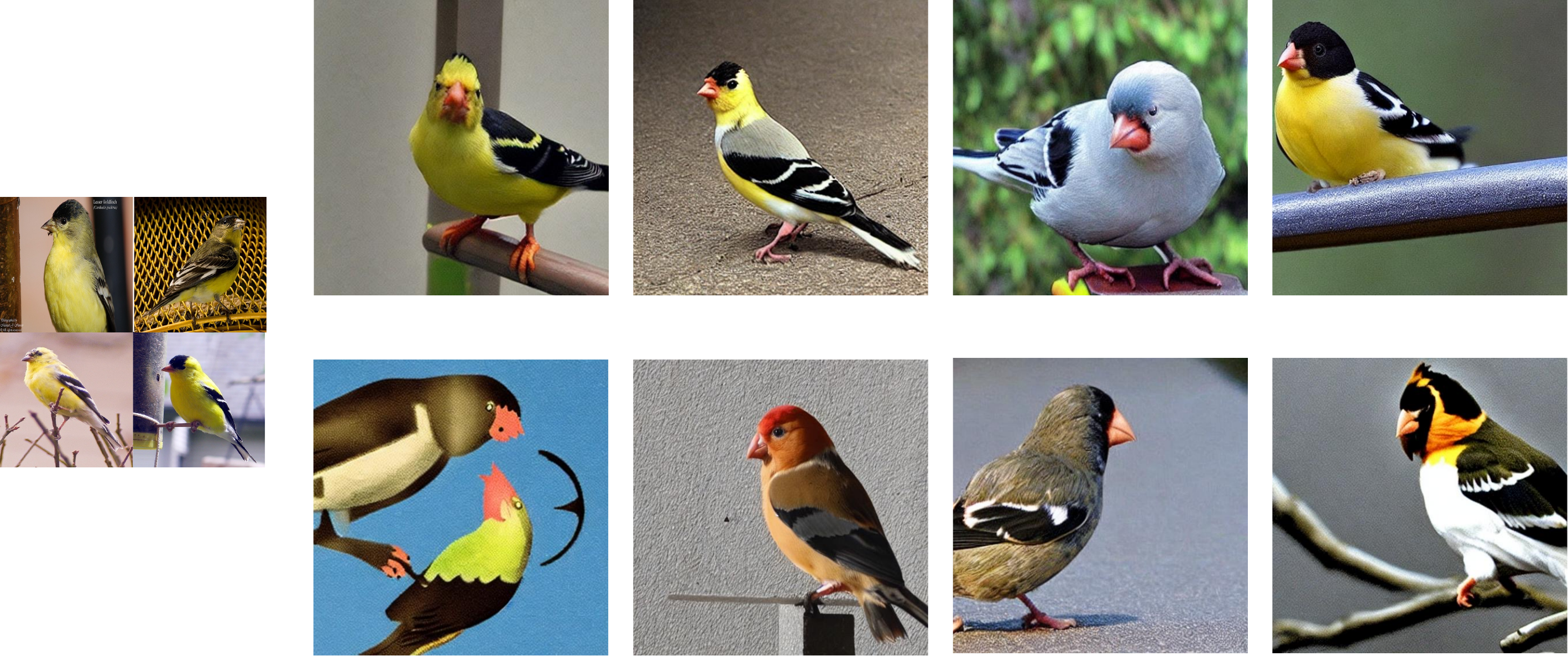}
    \caption{Qualitative examples from Custom Diffusions at different random seeds. The leftmost four figures are the target concept images. Top-Right four images are object-specific generated images. While Bottom-Right four generated images are on different composite text prompts.  }
    \label{fig:qual_randomseed}
\end{figure*}

Figure~\ref{fig:qual_style} presents qualitative examples showcasing the performance of various baseline methods across different style concepts. 
Notably, the textual inversion methods demonstrate limitations in preserving object-specific features and accurately learning the desired style. 
Furthermore, both DreamBooth and Custom Diffusion exhibit challenges in effectively capturing and reproducing the intended styles.
In Figure~\ref{fig:qual_object}, we delve into the object-specific learned concepts obtained through the baseline methodologies. 
Notably, Custom Diffusion struggles in acquiring and comprehending new concepts, thus explaining its relatively lower performance in terms of concept alignment.
To gain further insights, Figure~\ref{fig:qual_randomseed} offers a comparison of the generated images using Custom Diffusion at different random seeds. 
The results indicate that Custom Diffusion successfully generates the learned concepts in three out of four instances. 
However, when tasked with generating concept-specific images based on composite text prompts, Custom Diffusion struggles to maintain fidelity to the learned concept.

To facilitate a more comprehensive understanding of the \conceptbed{} benchmark and its results, we have developed an online results explorer, which provides readers with a user-friendly interface for exploring and analyzing the benchmark outcomes.

\section{Limitations}
We introduce the first comprehensive benchmark for large-scale concept learning, encompassing 284 distinct concepts and a vast collection of 33,000 composite prompts.
However, there are infinitely many concepts, and evaluating all of them is next to impossible.
Therefore, we recommend that future works benchmark the novel methodologies with the combination of both \conceptbed{} and selective qualitative examples.
While training and evaluating numerous models on an expanded subset of concepts can be resource-intensive, our approach, \conceptbed{}, employs an automated strategy that effortlessly scales to incorporate an extensive range of concepts.
Our benchmark primarily evaluates concept learning strategies derived from Stable Diffusion models. 
However, the dataset and evaluation framework we present in \conceptbed{} can serve as a good foundation for assessing any text-conditioned concept learners, including inversion methodologies.
It is important to note that the limitations inherent to Stable Diffusion models, which form the core of our experiments, extend to other concept learners, such as spatial relationships. 
Hence, while \conceptbed{} utilizes composite text prompts pre-trained on text-to-image models, future work will explore strategies to enable concept learners to adapt rapidly to novel concepts and achieve state-of-the-art performance on our benchmark.
In addition to the above, concept learning holds promise for enhancing performance in various application domains, such as refining existing concepts to mitigate potential biases present in Stable Diffusion models and incorporating spatial relations like left/right. 
These areas offer fertile ground for further exploration and can contribute to the advancement of concept learning techniques.
By addressing these limitations and exploring potential application areas, we aim to propel the development of concept learning methods that consistently push the boundaries of performance on the \conceptbed{} benchmark.

\begin{table}[!t]
    \centering
    \small
    \resizebox{\linewidth}{!}{%
    
    \begin{tabular}{@{}l cccc@{}}
        \toprule
        \textbf{Models} &   \textbf{ResNet18} &       \textbf{Inception-V4} &       \textbf{ViT-Large} &     \textbf{ConvNeXt} \\ %& \textbf{Human Score} \\
        \midrule
        Textual Inversion (LDM)              & 0.0107 & 0.0773 & 0.1165 & 0.0955\\ % & 4.069 \\
        Textual Inversion (SD)    & -0.0100 & 0.0201 & 0.0599 & 0.0472 \\ %& 4.159 \\
        DreamBooth & 0.0214 & 0.0485 & 0.0786  & 0.0678 \\ %& 4.075 \\
        Custom Diffusion      &  0.1538 & 0.1845 & 0.2286 & 0.2085 \\ %& 3.803 \\
        \midrule
        original                          & 0.0000 & 0.0000 & 0.0000 &  0.0000 \\ %& - \\
        
        \bottomrule
        
    \end{tabular}%
    }
    \caption{\textbf{Effects of different classifiers as Oracles.} This table summarizes the $\metric$ performance based on the different types of classifiers across the parameters range.}
    \label{tab:diff_cls}
\end{table}

\clearpage
\begin{table*}
    \centering
    
    \resizebox{0.7\linewidth}{!}{%
    \begin{tabular}{@{}c L@{}}
        \toprule
        \textbf{Concept Source} & \textbf{Concepts} \\
        \midrule 
        PACS & { \tiny \concept{Art-Painting} \quad \concept{Cartoon} \quad \concept{Photo} \quad \concept{Sketch} }\\ % \multicolumn{1}{m{3cm}}{multi-line piece of text to showcase a multi-line and justified cell} 
        \midrule
        ImageNet & { \tiny \concept{langur} \quad \concept{hand-held\_computer} \quad \concept{guenon} \quad \concept{brambling} \quad \concept{desktop\_computer} \quad \concept{speedboat} \quad \concept{titi} \quad \concept{airship} \quad \concept{tiger\_cat} \quad \concept{organ} \quad \concept{squirrel\_monkey} \quad \concept{bluetick} \quad \concept{siamang} \quad \concept{yawl} \quad \concept{lifeboat} \quad \concept{ambulance} \quad \concept{beagle} \quad \concept{digital\_clock} \quad \concept{fire\_engine} \quad \concept{Walker\_hound} \quad \concept{gondola} \quad \concept{pill\_bottle} \quad \concept{fireboat} \quad \concept{proboscis\_monkey} \quad \concept{moving\_van} \quad \concept{rotisserie} \quad \concept{slide\_rule} \quad \concept{Irish\_wolfhound} \quad \concept{junco} \quad \concept{cab} \quad \concept{magpie} \quad \concept{robin} \quad \concept{jeep} \quad \concept{colobus} \quad \concept{airliner} \quad \concept{gibbon} \quad \concept{letter\_opener} \quad \concept{garbage\_truck} \quad \concept{limousine} \quad \concept{English\_foxhound} \quad \concept{borzoi} \quad \concept{baboon} \quad \concept{basset} \quad \concept{capuchin} \quad \concept{convertible} \quad \concept{analog\_clock} \quad \concept{redbone} \quad \concept{canoe} \quad \concept{spider\_monkey} \quad \concept{bulbul} \quad \concept{Afghan\_hound} \quad \concept{goldfinch} \quad \concept{patas} \quad \concept{tabby} \quad \concept{web\_site} \quad \concept{grand\_piano} \quad \concept{laptop} \quad \concept{chickadee} \quad \concept{Dutch\_oven} \quad \concept{black-and-tan\_coonhound} \quad \concept{marmoset} \quad \concept{chimpanzee} \quad \concept{macaque} \quad \concept{police\_van} \quad \concept{tow\_truck} \quad \concept{cleaver} \quad \concept{howler\_monkey} \quad \concept{bloodhound} \quad \concept{pickup} \quad \concept{house\_finch} \quad \concept{beer\_bottle} \quad \concept{notebook} \quad \concept{water\_ouzel} \quad \concept{orangutan} \quad \concept{Madagascar\_cat} \quad \concept{gorilla} \quad \concept{indri} \quad \concept{beach\_wagon} \quad \concept{jay} \quad \concept{indigo\_bunting} }\\
        \midrule
        CUB & { \tiny \concept{Black\_footed\_Albatross} \quad \concept{Laysan\_Albatross} \quad \concept{Sooty\_Albatross} \quad \concept{Groove\_billed\_Ani} \quad \concept{Crested\_Auklet} \quad \concept{Least\_Auklet} \quad \concept{Parakeet\_Auklet} \quad \concept{Rhinoceros\_Auklet} \quad \concept{Brewer\_Blackbird} \quad \concept{Red\_winged\_Blackbird} \quad \concept{Rusty\_Blackbird} \quad \concept{Yellow\_headed\_Blackbird} \quad \concept{Bobolink} \quad \concept{Indigo\_Bunting} \quad \concept{Lazuli\_Bunting} \quad \concept{Painted\_Bunting} \quad \concept{Cardinal} \quad \concept{Spotted\_Catbird} \quad \concept{Gray\_Catbird} \quad \concept{Yellow\_breasted\_Chat} \quad \concept{Eastern\_Towhee} \quad \concept{Chuck\_will\_Widow} \quad \concept{Brandt\_Cormorant} \quad \concept{Red\_faced\_Cormorant} \quad \concept{Pelagic\_Cormorant} \quad \concept{Bronzed\_Cowbird} \quad \concept{Shiny\_Cowbird} \quad \concept{Brown\_Creeper} \quad \concept{American\_Crow} \quad \concept{Fish\_Crow} \quad \concept{Black\_billed\_Cuckoo} \quad \concept{Mangrove\_Cuckoo} \quad \concept{Yellow\_billed\_Cuckoo} \quad \concept{Gray\_crowned\_Rosy\_Finch} \quad \concept{Purple\_Finch} \quad \concept{Northern\_Flicker} \quad \concept{Acadian\_Flycatcher} \quad \concept{Great\_Crested\_Flycatcher} \quad \concept{Least\_Flycatcher} \quad \concept{Olive\_sided\_Flycatcher} \quad \concept{Scissor\_tailed\_Flycatcher} \quad \concept{Vermilion\_Flycatcher} \quad \concept{Yellow\_bellied\_Flycatcher} \quad \concept{Frigatebird} \quad \concept{Northern\_Fulmar} \quad \concept{Gadwall} \quad \concept{American\_Goldfinch} \quad \concept{European\_Goldfinch} \quad \concept{Boat\_tailed\_Grackle} \quad \concept{Eared\_Grebe} \quad \concept{Horned\_Grebe} \quad \concept{Pied\_billed\_Grebe} \quad \concept{Western\_Grebe} \quad \concept{Blue\_Grosbeak} \quad \concept{Evening\_Grosbeak} \quad \concept{Pine\_Grosbeak} \quad \concept{Rose\_breasted\_Grosbeak} \quad \concept{Pigeon\_Guillemot} \quad \concept{California\_Gull} \quad \concept{Glaucous\_winged\_Gull} \quad \concept{Heermann\_Gull} \quad \concept{Herring\_Gull} \quad \concept{Ivory\_Gull} \quad \concept{Ring\_billed\_Gull} \quad \concept{Slaty\_backed\_Gull} \quad \concept{Western\_Gull} \quad \concept{Anna\_Hummingbird} \quad \concept{Ruby\_throated\_Hummingbird} \quad \concept{Rufous\_Hummingbird} \quad \concept{Green\_Violetear} \quad \concept{Long\_tailed\_Jaeger} \quad \concept{Pomarine\_Jaeger} \quad \concept{Blue\_Jay} \quad \concept{Florida\_Jay} \quad \concept{Green\_Jay} \quad \concept{Dark\_eyed\_Junco} \quad \concept{Tropical\_Kingbird} \quad \concept{Gray\_Kingbird} \quad \concept{Belted\_Kingfisher} \quad \concept{Green\_Kingfisher} \quad \concept{Pied\_Kingfisher} \quad \concept{Ringed\_Kingfisher} \quad \concept{White\_breasted\_Kingfisher} \quad \concept{Red\_legged\_Kittiwake} \quad \concept{Horned\_Lark} \quad \concept{Pacific\_Loon} \quad \concept{Mallard} \quad \concept{Western\_Meadowlark} \quad \concept{Hooded\_Merganser} \quad \concept{Red\_breasted\_Merganser} \quad \concept{Mockingbird} \quad \concept{Nighthawk} \quad \concept{Clark\_Nutcracker} \quad \concept{White\_breasted\_Nuthatch} \quad \concept{Baltimore\_Oriole} \quad \concept{Hooded\_Oriole} \quad \concept{Orchard\_Oriole} \quad \concept{Scott\_Oriole} \quad \concept{Ovenbird} \quad \concept{Brown\_Pelican} \quad \concept{White\_Pelican} \quad \concept{Western\_Wood\_Pewee} \quad \concept{Sayornis} \quad \concept{American\_Pipit} \quad \concept{Whip\_poor\_Will} \quad \concept{Horned\_Puffin} \quad \concept{Common\_Raven} \quad \concept{White\_necked\_Raven} \quad \concept{American\_Redstart} \quad \concept{Geococcyx} \quad \concept{Loggerhead\_Shrike} \quad \concept{Great\_Grey\_Shrike} \quad \concept{Baird\_Sparrow} \quad \concept{Black\_throated\_Sparrow} \quad \concept{Brewer\_Sparrow} \quad \concept{Chipping\_Sparrow} \quad \concept{Clay\_colored\_Sparrow} \quad \concept{House\_Sparrow} \quad \concept{Field\_Sparrow} \quad \concept{Fox\_Sparrow} \quad \concept{Grasshopper\_Sparrow} \quad \concept{Harris\_Sparrow} \quad \concept{Henslow\_Sparrow} \quad \concept{Le\_Conte\_Sparrow} \quad \concept{Lincoln\_Sparrow} \quad \concept{Nelson\_Sharp\_tailed\_Sparrow} \quad \concept{Savannah\_Sparrow} \quad \concept{Seaside\_Sparrow} \quad \concept{Song\_Sparrow} \quad \concept{Tree\_Sparrow} \quad \concept{Vesper\_Sparrow} \quad \concept{White\_crowned\_Sparrow} \quad \concept{White\_throated\_Sparrow} \quad \concept{Cape\_Glossy\_Starling} \quad \concept{Bank\_Swallow} \quad \concept{Barn\_Swallow} \quad \concept{Cliff\_Swallow} \quad \concept{Tree\_Swallow} \quad \concept{Scarlet\_Tanager} \quad \concept{Summer\_Tanager} \quad \concept{Artic\_Tern} \quad \concept{Black\_Tern} \quad \concept{Caspian\_Tern} \quad \concept{Common\_Tern} \quad \concept{Elegant\_Tern} \quad \concept{Forsters\_Tern} \quad \concept{Least\_Tern} \quad \concept{Green\_tailed\_Towhee} \quad \concept{Brown\_Thrasher} \quad \concept{Sage\_Thrasher} \quad \concept{Black\_capped\_Vireo} \quad \concept{Blue\_headed\_Vireo} \quad \concept{Philadelphia\_Vireo} \quad \concept{Red\_eyed\_Vireo} \quad \concept{Warbling\_Vireo} \quad \concept{White\_eyed\_Vireo} \quad \concept{Yellow\_throated\_Vireo} \quad \concept{Bay\_breasted\_Warbler} \quad \concept{Black\_and\_white\_Warbler} \quad \concept{Black\_throated\_Blue\_Warbler} \quad \concept{Blue\_winged\_Warbler} \quad \concept{Canada\_Warbler} \quad \concept{Cape\_May\_Warbler} \quad \concept{Cerulean\_Warbler} \quad \concept{Chestnut\_sided\_Warbler} \quad \concept{Golden\_winged\_Warbler} \quad \concept{Hooded\_Warbler} \quad \concept{Kentucky\_Warbler} \quad \concept{Magnolia\_Warbler} \quad \concept{Mourning\_Warbler} \quad \concept{Myrtle\_Warbler} \quad \concept{Nashville\_Warbler} \quad \concept{Orange\_crowned\_Warbler} \quad \concept{Palm\_Warbler} \quad \concept{Pine\_Warbler} \quad \concept{Prairie\_Warbler} \quad \concept{Prothonotary\_Warbler} \quad \concept{Swainson\_Warbler} \quad \concept{Tennessee\_Warbler} \quad \concept{Wilson\_Warbler} \quad \concept{Worm\_eating\_Warbler} \quad \concept{Yellow\_Warbler} \quad \concept{Northern\_Waterthrush} \quad \concept{Louisiana\_Waterthrush} \quad \concept{Bohemian\_Waxwing} \quad \concept{Cedar\_Waxwing} \quad \concept{American\_Three\_toed\_Woodpecker} \quad \concept{Pileated\_Woodpecker} \quad \concept{Red\_bellied\_Woodpecker} \quad \concept{Red\_cockaded\_Woodpecker} \quad \concept{Red\_headed\_Woodpecker} \quad \concept{Downy\_Woodpecker} \quad \concept{Bewick\_Wren} \quad \concept{Cactus\_Wren} \quad \concept{Carolina\_Wren} \quad \concept{House\_Wren} \quad \concept{Marsh\_Wren} \quad \concept{Rock\_Wren} \quad \concept{Winter\_Wren} \quad \concept{Common\_Yellowthroat} } \\
        \bottomrule
    \end{tabular}%
    }
    \caption{List of all concepts from \conceptbed{} library based on their data source.}
    \label{tab:concepts}
\end{table*}
\begin{table*}[ht]
    \centering
    \tiny
    % \resizebox{\textwidth}{!}{%
    
    \begin{tabular}{@{}l ccccc@{}}
        \toprule
        \textbf{Concept} & \textbf{Action} & \textbf{Attribute} & \textbf{Counting} & \textbf{Relation} & \textbf{Overall} \\
        \midrule
                 \concept{laptop} &      17 &         18 &         2 &        40 &       52 \\
              \concept{tow\_truck} &      97 &        348 &        35 &       409 &      645 \\
     \concept{hand-held\_computer} &      17 &         18 &         2 &        40 &       52 \\
                \concept{gorilla} &       9 &         11 &         0 &        11 &       19 \\
             \concept{chimpanzee} &       9 &         11 &         0 &        11 &       19 \\
                 \concept{pickup} &      97 &        348 &        35 &       409 &      645 \\
                   \concept{yawl} &     116 &        178 &        36 &       466 &      567 \\
                 \concept{beagle} &     380 &        807 &        24 &       496 &     1222 \\
                 \concept{bulbul} &      62 &        270 &        11 &       142 &      374 \\
          \concept{spider\_monkey} &       9 &         11 &         0 &        11 &       19 \\
                 \concept{borzoi} &     380 &        807 &        24 &       496 &     1222 \\
           \concept{analog\_clock} &       1 &         61 &        10 &        71 &      109 \\
          \concept{letter\_opener} &      11 &         10 &         0 &        21 &       31 \\
            \concept{water\_ouzel} &      62 &        270 &        11 &       142 &      374 \\
               \concept{web\_site} &      17 &         18 &         2 &        40 &       52 \\
          \concept{garbage\_truck} &      97 &        348 &        35 &       409 &      645 \\
             \concept{bloodhound} &     380 &        807 &        24 &       496 &     1222 \\
                 \concept{basset} &     380 &        807 &        24 &       496 &     1222 \\
       \concept{proboscis\_monkey} &       9 &         11 &         0 &        11 &       19 \\
             \concept{Dutch\_oven} &      58 &         85 &        13 &       111 &      194 \\
               \concept{fireboat} &     116 &        178 &        36 &       466 &      567 \\
\concept{black-and-tan\_coonhound} &     380 &        807 &        24 &       496 &     1222 \\
              \concept{speedboat} &     116 &        178 &        36 &       466 &      567 \\
            \concept{beach\_wagon} &      98 &        213 &        17 &       363 &      497 \\
               \concept{airliner} &       4 &          8 &         2 &        11 &       20 \\
                   \concept{titi} &       9 &         11 &         0 &        11 &       19 \\
               \concept{marmoset} &       9 &         11 &         0 &        11 &       19 \\
            \concept{beer\_bottle} &       1 &          9 &         0 &        14 &       20 \\
                 \concept{magpie} &      62 &        270 &        11 &       142 &      374 \\
        \concept{Irish\_wolfhound} &     380 &        807 &        24 &       496 &     1222 \\
               \concept{lifeboat} &     116 &        178 &        36 &       466 &      567 \\
              \concept{brambling} &      62 &        270 &        11 &       142 &      374 \\
             \concept{rotisserie} &      58 &         85 &        13 &       111 &      194 \\
                  \concept{junco} &      62 &        270 &        11 &       142 &      374 \\
              \concept{ambulance} &      98 &        213 &        17 &       363 &      497 \\
                \concept{gondola} &     116 &        178 &        36 &       466 &      567 \\
                  \concept{tabby} &     424 &        992 &        42 &       727 &     1592 \\
                \concept{cleaver} &      11 &         10 &         0 &        21 &       31 \\
              \concept{limousine} &      98 &        213 &        17 &       363 &      497 \\
       \concept{desktop\_computer} &      17 &         18 &         2 &        40 &       52 \\
                \concept{colobus} &       9 &         11 &         0 &        11 &       19 \\
            \concept{house\_finch} &      62 &        270 &        11 &       142 &      374 \\
              \concept{chickadee} &      62 &        270 &        11 &       142 &      374 \\
                    \concept{cab} &      98 &        213 &        17 &       363 &      497 \\
               \concept{notebook} &      17 &         18 &         2 &        40 &       52 \\
        \concept{squirrel\_monkey} &       9 &         11 &         0 &        11 &       19 \\
          \concept{digital\_clock} &       1 &         61 &        10 &        71 &      109 \\
                  \concept{canoe} &     116 &        178 &        36 &       466 &      567 \\
                  \concept{indri} &       9 &         11 &         0 &        11 &       19 \\
       \concept{English\_foxhound} &     380 &        807 &        24 &       496 &     1222 \\
                \concept{airship} &       4 &          8 &         2 &        11 &       20 \\
               \concept{capuchin} &       9 &         11 &         0 &        11 &       19 \\
              \concept{tiger\_cat} &     424 &        992 &        42 &       727 &     1592 \\
               \concept{bluetick} &     380 &        807 &        24 &       496 &     1222 \\
           \concept{Afghan\_hound} &     380 &        807 &        24 &       496 &     1222 \\
             \concept{moving\_van} &      97 &        348 &        35 &       409 &      645 \\
                    \concept{jay} &      62 &        270 &        11 &       142 &      374 \\
             \concept{police\_van} &      97 &        348 &        35 &       409 &      645 \\
          \concept{howler\_monkey} &       9 &         11 &         0 &        11 &       19 \\
                 \concept{langur} &       9 &         11 &         0 &        11 &       19 \\
                 \concept{gibbon} &       9 &         11 &         0 &        11 &       19 \\
                \concept{redbone} &     380 &        807 &        24 &       496 &     1222 \\
                  \concept{organ} &       3 &         24 &        12 &        41 &       68 \\
             \concept{slide\_rule} &      17 &         18 &         2 &        40 &       52 \\
              \concept{goldfinch} &      62 &        270 &        11 &       142 &      374 \\
            \concept{pill\_bottle} &       1 &          9 &         0 &        14 &       20 \\
                \concept{siamang} &       9 &         11 &         0 &        11 &       19 \\
            \concept{convertible} &      98 &        213 &        17 &       363 &      497 \\
                 \concept{baboon} &       9 &         11 &         0 &        11 &       19 \\
           \concept{Walker\_hound} &     380 &        807 &        24 &       496 &     1222 \\
                 \concept{guenon} &       9 &         11 &         0 &        11 &       19 \\
         \concept{indigo\_bunting} &      62 &        270 &        11 &       142 &      374 \\
            \concept{grand\_piano} &       3 &         24 &        12 &        41 &       68 \\
            \concept{fire\_engine} &      97 &        348 &        35 &       409 &      645 \\
                  \concept{robin} &      62 &        270 &        11 &       142 &      374 \\
                \concept{macaque} &       9 &         11 &         0 &        11 &       19 \\
              \concept{orangutan} &       9 &         11 &         0 &        11 &       19 \\
                   \concept{jeep} &      98 &        213 &        17 &       363 &      497 \\
                  \concept{patas} &       9 &         11 &         0 &        11 &       19 \\
         \concept{Madagascar\_cat} &       9 &         11 &         0 &        11 &       19 \\

        \bottomrule
        
    \end{tabular}%
    % }
    \caption{This table shows the composition statistics by categories. Here, overall means the unique compositions per concept and less than or equal to the sum of all four compositions as one composite prompt can belong up to two composition categories.}
    \label{tab:concept_stats}
\end{table*}
\clearpage

\clearpage

% \clearpage

% \setcitestyle{numbers,square}
% \bibliographystyle{abbrv}
% \bibliography{llc}

% The ability to understand visual concepts and replicate and compose these concepts from images is a central goal for computer vision. Recent advances in text-to-image (T2I) models have lead to high definition and realistic image quality generation by learning from large databases of images and their descriptions. However, the evaluation of T2I models has focused on photorealism and limited qualitative measures of visual understanding. To quantify the ability of T2I models in learning and synthesizing novel visual concepts (a.k.a. personalized T2I), we introduce ConceptBed, a large-scale dataset that consists of 284 unique visual concepts, and 33K composite text prompts. Along with the dataset, we propose an evaluation metric, Concept Confidence Deviation (CCD), that uses the confidence of oracle concept classifiers to measure the alignment between concepts generated by T2I generators and concepts contained in target images. We evaluate visual concepts that are either objects, attributes, or styles, and also evaluate four dimensions of compositionality: counting, attributes, relations, and actions. Our human study shows that CCD is highly correlated with human understanding of concepts. Our results point to a trade-off between learning the concepts and preserving the compositionality which existing approaches struggle to overcome. The data, code, and interactive demo is available at: https://conceptbed.github.io/

\end{document}